%% file: main_arxiv.tex
\documentclass{article}
\usepackage{PRIMEarxiv}
\usepackage{subcaption}
\usepackage{float}
\usepackage{caption} 
\usepackage{placeins} 
\usepackage[utf8]{inputenc} 
\usepackage[T1]{fontenc}    
\usepackage{hyperref}       
\usepackage{url}            
\usepackage{booktabs}       
\usepackage{amsfonts}       
\usepackage{nicefrac}       
\usepackage{microtype}      
\usepackage{lipsum}
\usepackage{fancyhdr}       
\usepackage{graphicx}       
\usepackage{adjustbox}
\usepackage{amsmath,amssymb,amsfonts}
\usepackage{algorithmic}
\usepackage{graphicx}
\usepackage{booktabs}
\usepackage{textcomp}
\usepackage{enumerate}
\usepackage{amscd}
\newtheorem{theorem}{Theorem}

\usepackage[ruled,linesnumbered]{algorithm2e}
\usepackage[nolist, nohyperlinks, printonlyused]{acronym} 
\usepackage{epsfig} 
\usepackage{epstopdf}
\usepackage{verbatim}
\usepackage{array}
\usepackage{caption}
\usepackage{amssymb}
\usepackage{fancyhdr}
\usepackage{enumitem}
\usepackage{tabularx}
\graphicspath{{media/}}  
\usepackage{indentfirst} 
\usepackage{multirow}
\usepackage{tabularray}
\title{GMapLatent: Geometric Mapping in Latent Space
\thanks{\textit{\underline{Citation}}: 
\textbf{Authors. Title. Pages.... DOI:000000/11111.}} 
}

\author{
 Wei Zeng\thanks{Corresponding author: wz@xjtu.edu.cn}, Xuebin Chang, Jianghao Su, Xiang Gu, Jian Sun, Zongben Xu 
\\
  Dept. of Information Science, School of Mathematics and Statistics \\
   Xi'an Jiaotong University, 
  Xi'an, Shaanxi, China\\
}

\begin{document}
\maketitle

\input{0_abstract}




\input{1_introduction}
\input{2_related_work}
\input{3_background}

\input{4_algorithm}

\input{5_model}
\input{6_experiment}

\input{7_conclusion}

\section*{Acknowledgments}
This work was supported by the National Key R\&D Program 2021YFA1003002 and the National Natural Science Foundation of China (12090021 and 12090020).

\bibliographystyle{unsrt}  
\bibliography{references}

\end{document}

%% file: 0_abstract.tex
\begin{abstract}
Cross-domain generative models based on encoder-decoder AI architectures have attracted much attention in generating realistic images, where domain alignment is crucial for generation accuracy. 
Domain alignment methods usually deal directly with the initial distribution; however,
mismatched or mixed clusters can lead to mode collapse and mixture problems in the decoder, compromising model generalization capabilities.
In this work, we innovate a cross-domain alignment and generation model that introduces a canonical latent space representation based on geometric mapping to align the cross-domain latent spaces in a rigorous and precise manner, thus avoiding mode collapse and mixture in the encoder-decoder generation architectures. We name this model GMapLatent. 
The core of the method is to seamlessly align latent spaces with strict cluster correspondence constraints using the canonical parameterizations of cluster-decorated latent spaces. We first (1) transform the latent space to a canonical parameter domain by composing barycenter translation, optimal transport merging and constrained harmonic mapping, and then (2) compute geometric registration with cluster constraints over the canonical parameter domains. This process realizes a bijective (one-to-one and onto) mapping between newly transformed latent spaces and generates a precise alignment of cluster pairs. 
Cross-domain generation is then achieved through the aligned latent spaces embedded in the encoder-decoder pipeline. 
Experiments on gray-scale and color images validate the efficiency, efficacy and applicability of GMapLatent, and demonstrate that the proposed model has superior performance over existing models.  
\end{abstract}

%% file: 1_introduction.tex
\section{Introduction}
\label{sec:introduction}

\begin{figure}[ht]
	\centering
	\hspace{2mm}\includegraphics[width=0.45\textwidth]{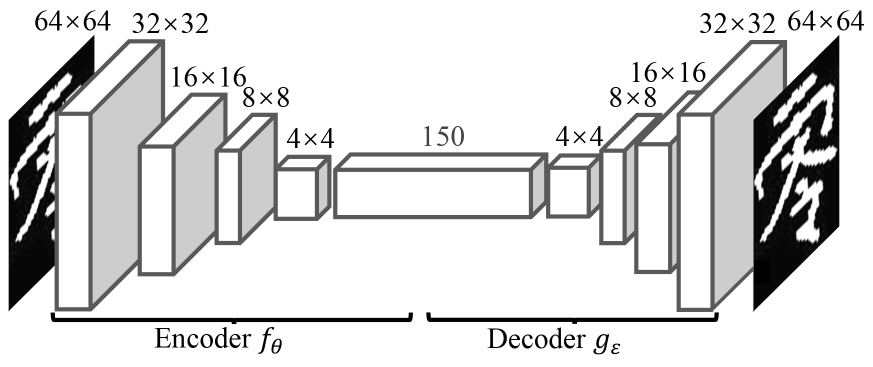}
	\caption{Autoencoder architecture for Chinese MNIST dataset. $f_\theta$ and $g_\varepsilon$ represent the encoding and decoding maps, respectively, where $\theta$ and $\varepsilon$ are their corresponding network parameters.}
        \label{fig:AE}
\end{figure}

Generating from one modality to another can be realized by cross-domain generative models based on deep networks \cite{liu2017unsupervised}. A feasible strategy is to insert the latent code conversion module to connect the latent spaces constructed in the encoder-decoder architectures of source and target domains.
For generality, we work on multi-class cross-domain generation problem. Given a dataset with multiple classes, the ambient space can be embedded into the corresponding latent space with the techniques of encoding (e.g., autoencoder (AE) \cite{sainath2012auto}, see Fig. \ref{fig:AE}) and dimensionality reduction  (e.g., UMAP \cite{mcinnes2018umap} or t-SNE \cite{van2008visualizing}), which project high-dimensional data to low-dimensional (typically, 2D or 3D) space. The latent space for a real-world dataset always contains outliers and class mixtures (see Fig. \ref{fig:latent_space} for an example). This is mainly due to the confusion and mixture of data samples or insufficient distinguishing ability of data embedding techniques. And, different datasets may have different cluster distribution configurations (geometries and topologies) in their latent spaces, i.e., different cluster locations, sizes and patterns. 
In the framework of latent space-based cross-domain generation, such vast differences pose a serious challenge for latent code transformation from source to target. 
Then, precise multi-class cross-domain generation refers to the establishing alignment between cross-domain latent spaces with hard cluster-to-cluster constraints. 

This work aims to build a cross-domain alignment and generation model by accurately and seamlessly aligning 2D latent spaces of source and target ambient spaces, where latent spaces are transformed and the correspondence between semantic clusters are strictly constrained. 
The final generation is essentially a diffeomorphism between the transformed source and target latent spaces, which is expected to be a one-to-one, onto and continuous mapping to avoid mode collapse or mixture in translation tasks. 
Existing alignment methods across domains can achieve point-to-point generation across different classes \cite{zhu2017unpaired, gu2022keypoint}, but has no ability to address mode collapse and mixture and curve-to-curve continuous generation. 
This study endeavors to overcome the inherent limitations of direct mapping of discrete sample points and liberate the generation process with the help of geometric mapping methods, so as to avoid mode collapse and mixture and accomplish continuous generation from curve to curve.

{\setlength{\tabcolsep}{1pt} 
\begin{figure}[ht]
	\centering
        \footnotesize
 \begin{tabular}{c}
\includegraphics[height=0.18\textwidth]{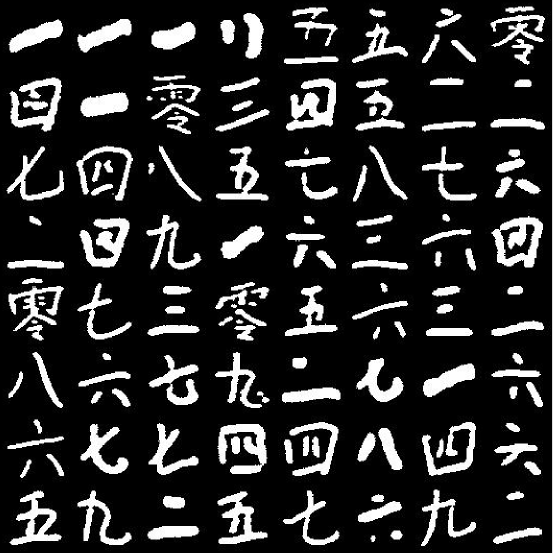}
\includegraphics[height=0.18\textwidth]{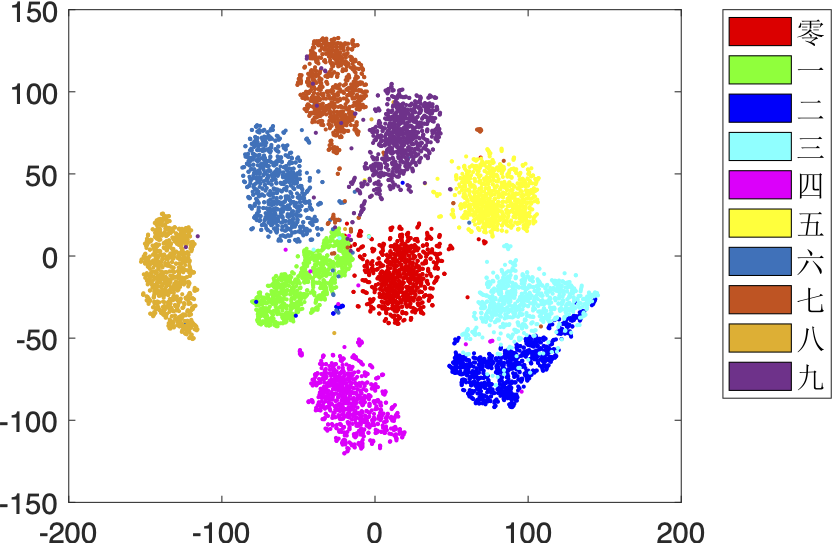}\\    
(a) Chinese MNIST \\
\includegraphics[height=0.18\textwidth]{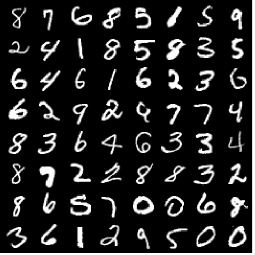}
\includegraphics[height=0.18\textwidth]{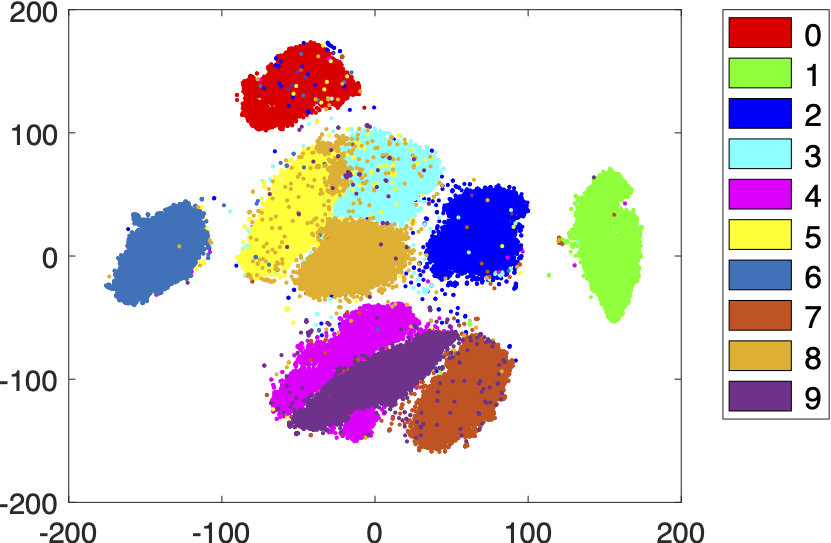}\\
(b) Arabic MNIST
 \end{tabular}
\caption{T-SNE embedding results for two handwritten digit datasets \cite{ChineseMnist,deng2012mnist}. 
 }
\label{fig:latent_space}
\end{figure}
}

\subsection{Motivation}
The motivation for this work is to incorporate differential geometric mappings into latent space, to study representations of latent space and mappings between latent spaces, and to extend the methodology of studying geometric structures and mappings of surfaces to study those of latent spaces. The resulting representations and mappings will then create novel cross-domain (covering single-domain) generative models.

\vspace{1mm}
\paragraph{Representation of latent space} Through learning representation techniques, data is encoded into a latent space, usually visualized as a low-dimensional space, where the structure of the data is effectively represented. 
Latent clusters are assembled by semantic categories according to the application setting and are the main components of our study of latent space. Their geometry and topology in the latent space essentially affects upstream and downstream analysis tasks within the deep learning framework, such as data generation, knowledge transfer, and further recognition and classification. The interior of a cluster represents the distribution of corresponding semantic category, but may not appear uniform depending on the given data sampling and embedding technique. The boundaries of the clusters are an important feature that may lead to mode collapse or mixture problems. The gaps between clusters are those undiscovered modes outside the data manifold and  therefore are considered to have zero probability measure of being captured from the given data.

\begin{figure}[ht]
	\centering
	\includegraphics[width=.48\textwidth]{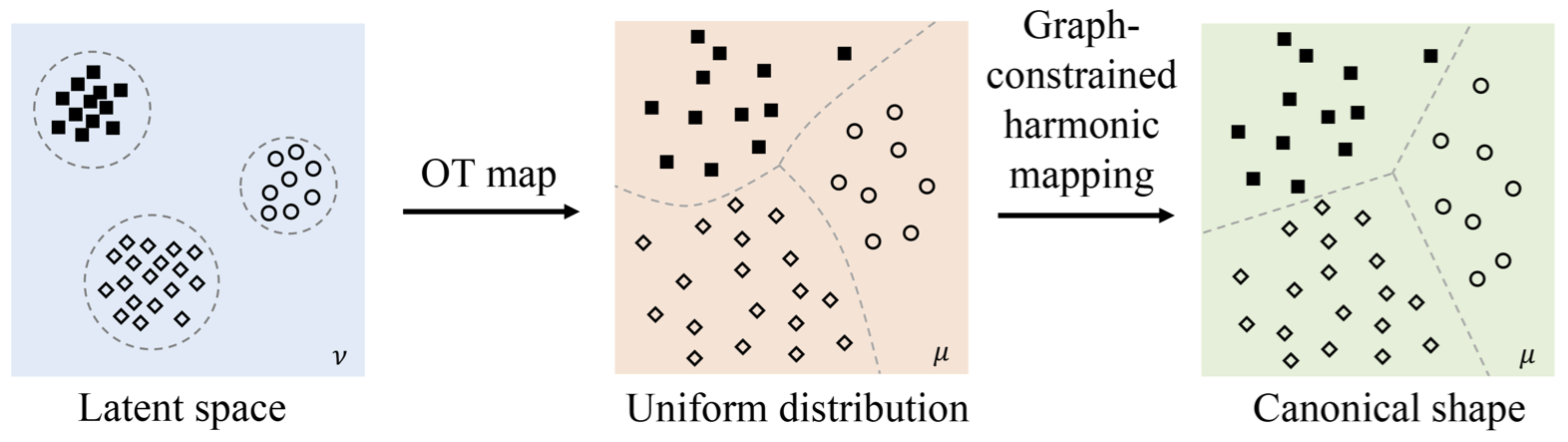}
	\caption{Canonical representation of latent space. 
    }
	\label{fig:Mapper}
\end{figure}

Based on the above understanding, as a foundational theory, optimal transport (OT) provides a way to seamlessly merge latent clusters \cite{an2019ae} considering that the gaps between clusters have zero probability measure. With this mapping on latent space, the transformed clusters are assembled into a connected domain with no internal overlap between clusters, and their connected cluster boundaries form a curvy graph that distinguishes the categories on the connected domain. \textit{We call it a graph-decorated latent space} (i.e., a connected domain decorated with a curvy graph on it). Based on this observation, the graph-constrained surface parameterization \cite{Zeng2015Graph} is applicable to this graph-decorated latent space and can transform it to a canonical convex-subdivision domain (called \textit{canonical latent space}). Figure \ref{fig:Mapper} illustrates the original, merged, and canonical representations of latent space, respectively. Now, we have a novel \textit{structural representation} of latent space (or data) that respects the original geometry of clusters. In the canonical latent space, a curve can result in continuous intermediate results in the single-domain decoder, and the straight-line boundaries of clusters make it easy to locate the category of a sample and ensure precise alignment of cluster boundaries.

\vspace{1mm}
\paragraph{Alignment of latent spaces} Looking at two latent spaces, they usually have different geometric and topological structures. An immediate question is: How to align them, or to be precise, how to register them with a bijective (one-to-one and onto) mapping? This is the key technique needed to address the task of cross-domain distribution alignment and transfer. To the best of our knowledge, no existing study has solved the fully registered alignment between latent spaces, especially through a differential geometric mapping viewpoint. 

Looking at graph-decorated latent space representations and their canonical forms, 
it is natural to think of using geometric mappings to register/align latent spaces, where latent clusters are constraints. Before that, we should ensure that the topology of latent space remains consistent with necessary adjustment. Thus, leveraging the classical 3D-to-2D surface registration framework \cite{Zeng2018Graph}, we can achieve correspondence between latent spaces, and this alignment has theoretical bijection guarantee, i.e., one-to-one and onto, or orientation-preserving mappings.   

In this work, we initiate to integrate diffeomorphic geometric mappings to latent space in generative models.

\subsection{Algorithm overview} 
\begin{figure}[!ht]
	\centering
	\includegraphics[width=0.7\linewidth]{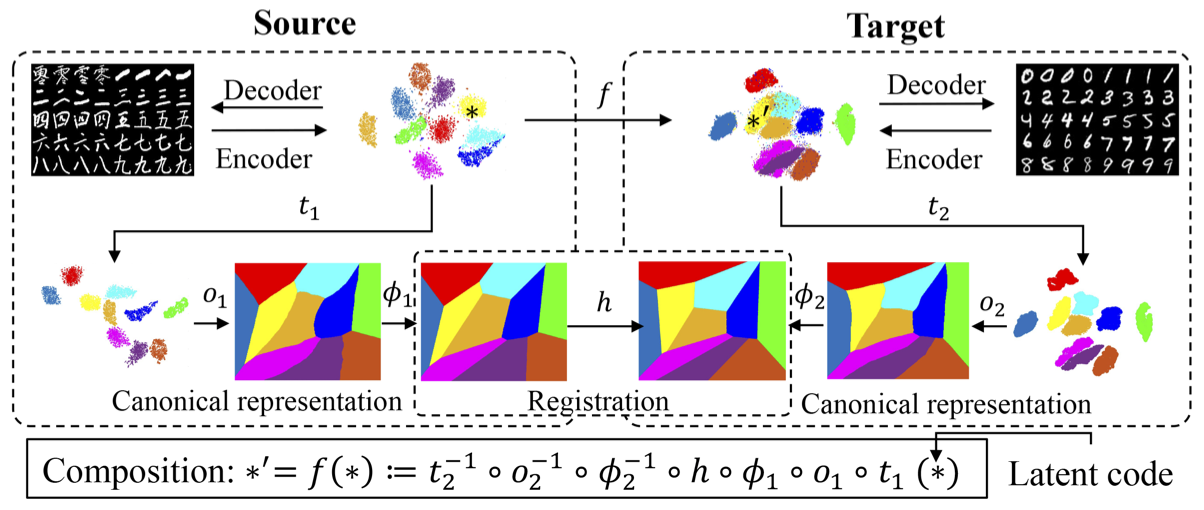}
	\caption{Workflow of generation from Chinese MNIST to Arabic MNIST. Given a latent code from source latent space, $f$ is the desired final mapping to generate the corresponding latent code in target latent space. $t_i$ - latent preprocessing (translation); $o_i$ - optimal transport merging; $\phi_i$ - canonical graph-constrained harmonic mapping (straightening); $h$ - graph-constrained harmonic registration (alignment).}
        \label{fig:cross}
\end{figure}

We present a novel framework to transform and register latent spaces via geometric mappings, and then embed a cross-domain registration module into the baseline encoder-decoder generation pipeline. Figure \ref{fig:cross} illustrates the workflow on image translation task. 
It computes the canonical shape representations for source and target latent spaces ($L_i, i=1,2$) and conduct the registration (alignment) over corresponding canonical parameter domains $D_i$'s. The final mapping $f:L_1 \to L_2$ is composed of a sequence of bijective transformations on latent clusters and latent spaces, intuitively, including preprocessing $t_i$, merging $o_i$, straightening $g_i$, their inverses, and alignment $h$, i.e., $f:=t_2^{-1} \circ o_2^{-1} \circ \phi_2^{-1} \circ h \circ \phi_1 \circ o_1 \circ t_1$. 

\textit{Cross-Domain Registration Module} is the key component of the entire generation process,  whose function is to perform a precise alignment of the source and target latent spaces. It includes the following four steps: 
(1) Perform barycentric translations on latent clusters to maintain topological consistency between latent spaces, establishing cluster correspondence constraints; 
(2) Compute optimal transport to merge latent clusters and remove intermediate margins in-between, in preparation for further diffeomorphic geometric processing; 
(3) Compute graph-constrained harmonic mappings to straighten the curvy graph-constrained domain to a canonical shape, i.e., a convex-subdivision domain, converting the curvy boundaries to straight lines. This generates a novel canonical shape representation of the latent space and converts nonlinear curve constraints to linear constraints in subsequent registration; and  
(4) Register the transformed latent spaces accurately and seamlessly, accomplishing strict alignment of both interior and boundaries of the corresponding latent clusters. 

\textit{Cross-Domain Generation Pipeline} works as follows: we apply an autoencoder \cite{sainath2012auto} to encode the image into a low-dimensional latent representation and further embed it to a 2-dimensional latent code by t-SNE \cite{van2008visualizing}. For the whole source and target datasets, we obtain the corresponding 2D latent spaces, as shown in Fig. \ref{fig:latent_space}. We then apply the above cross-domain registration operation to the 2D latent spaces and obtain a bijective mapping between them. Thus, given an image in the source space, we compute its latent code and find the corresponding latent code in the target latent space through the combinatorial processing of the geometric transformations described above, and then decode the projected latent code through the generator (decoder) of the target latent space to obtain the translation result. 

\subsection{Contributions} 
The novelty of this work can be summarized as follows:
\begin{enumerate}
    \item As we know, this is the first integration of \textit{diffeomorphic geometric mapping} to transform and register latent spaces in a seamless and continuous way. The proposed model is named GMapLatent; 
    \item It innovates an interpretable and controllable cross-domain generative model that avoids mode collapse and mixture by use of \textit{precise geometric registration} with \textit{latent cluster constraints}, and thus achieves curve-to-curve \textit{continuous cross-domain generation}; 
    \item It advances the field of data representation and generative AI modeling by treating \textit{latent space representation} from a differential geometry perspective and fusing geometric mappings in an effective way to solve the cross-domain generation problem;
    \item It achieves \textit{highly superior performance} over other existing models in cross-domain image translation tasks. The efficiency and applicability of GMapLatent are validated through comparison and ablation experiments. 
\end{enumerate}

The rest of the paper is organized as follows. Section \ref{sec:related_work} reviews the  related literature, and Section \ref{sec:background} describes the background theory and  methodologies. Section \ref{sec:algorithm} introduces our main algorithms and based on them Section \ref{sec:models} illustrates various strategies of constructing generative models. In Section \ref{sec:experiment}, we perform a comparative state-of-art evaluation for image translation tasks with ablation studies. Finally, Section \ref{sec:conclusion} concludes the paper with future work directions.

%% file: 2_related_work.tex
\section{Related works}
\label{sec:related_work}

Extensive research has been conducted on cross-domain alignment and generation models. Here, we will briefly review closely related works from the perspective of the structure and representation of latent spaces. 

\vspace{1mm}
\paragraph{Latent space representation and geometric generative model} The latent space of a multi-class dataset is composed of point clusters (see Fig. \ref{fig:latent_space}). The shape, size and position of clusters represent the geometry and topology of latent space. 
Recently, geometric approaches have been integrated into operating latent space. The study in GeoLatent \cite{yang2023geolatent} focuses on optimizing the latent space by designing a neural deformable shape generator that maps latent codes to 3D deformable shapes using a Riemannian metric of differential geometric deformation energies. 
Concurrently, the research in Glass \cite{muralikrishnan2022glass} addresses the challenge of training generative models on sparse datasets of 3D models and focuses on latent augmentation for 3D model generation and shape correspondence establishment. 
The AE-OT model \cite{an2019ae} is closely related to our work in terms of methodology, and it solves the mode collapse problem by using extended semi-discrete OT to generate continuous distributions in the latent space. Its goal is single-domain generation, rather than the cross-domain problem, where curvy cluster boundaries cause computational inefficiencies when localizing the class of a given sample. In addition, irregularly distributed and shaped clusters cannot be directly aligned across domains, which is the main problem we aim to address in this work. 
In summary, compared with our model, existing works do not introduce the concept of differential geometric mapping (e.g., harmonic mapping) into the latent space, nor do they explore the canonical structure and representation of latent space and the applicability of geometric mapping in cross-domain generative models. 

\vspace{1mm}
\paragraph{Cross-domain alignment} 
In this work, ``cross-domain'' alignment refers to building the correspondence between source and target domains, no matter whether they are in the same data space or not. In literature, there have been a large amount of works on domain adaptation 
\cite{farahani2021brief,kouw2019review} by aligning distributions in general feature space or latent space. 
Among them, for multi-class cross-domain alignment problem, there have been works which introduce constraints to enhance the accuracy of alignment. 
Gu et al. \cite{Gu2022KeypointGuidedOT} took key points as constraints to drive the alignment in optimal transport in generation framework. Yang et al.  \cite{yang2023prototypical} used prototypical constraints in optimal transport for universal domain adaptation to deal with the alignment of a labeled source domain to an unlabeled target domain. 
Note that these domain adaptation methods generate roughly aligned distributions in feature space or latent space. 
Specifically, in the case that semantic correspondences between classes are given, to the best of our knowledge, we could not find a work that can completely align all class cluster pairs (boundaries of clusters as constraints) by a whole one-to-one and onto mapping. Namely, cluster pairs (interior and boundaries) are fully registered \textit{in one shot}, and thus the generation has no mode collapse and mixture.

\vspace{1mm}
\paragraph{Geometric mapping} 
Geometric mapping of a single surface refers to surface parameterization, namely computing canonical parameter domain of a given surface, thus obtaining its shape representation. There have been conformal mapping \cite{Haker2000Conformal,Lvy2002LeastSC,gu2002computing,gu2007conformal,nehari2012conformal}, area-preserving mapping \cite{Zhu2003Area,Haker2004OptimalMT,zhao2013area}, harmonic mapping \cite{jost2006lectures}, quasiconformal mapping \cite{ahlfors2006lectures,zeng2009surface,zeng2012computing}, and mappings under various energy optimizations \cite{floater2005surface,Lui2010OptimizationOS}. Geometric mapping between surfaces refers to surface registration (dense alignment), namely building the bijective correspondence between surfaces, thus achieving the similarity metric for shape matching and comparison. 
Recently, geometric mapping of surfaces decorated with feature constraints (called decorated/constrained surfaces) has been studied \cite{CVPR14Point,ECCV14Curve,Zeng2015Graph,yang2024diffeomorphic}. In the research, canonical optimal quasiconformal mapping was proposed to study geometric structures of these surfaces and registrations between them with precise correspondence constraints. It targets solving practical tasks where physical constraints are commonly applied, e.g., facial registration with natural feature landmarks and curves. In our current work, 
the measure-preserving mapping is employed  to merge the clusters of 2D latent space and forms a connected domain, which is implemented by geometric optimal transport \cite{su2015optimal,an2019ae}. In addition to this, we specifically apply graph-constrained harmonic mapping to take cluster adjacency graphs as constraints to obtain the canonical convex-subdivision representation of latent space and build the complete correspondence between latent spaces, which is implemented by solving Laplacian equation \cite{Zeng2015Graph,Zeng2018Graph}.  

%% file: 3_background.tex
\section{Preliminaries}
\label{sec:background}
This work involves transforming latent space on both distribution and geometry, which correspond to optimal transport  and geometric mapping, respectively. 

\subsection{Optimal Transport}
\label{sec:background-OT}
Let $\Omega$ and $\Omega^{*}$ be two domains in Euclidean space $\mathbb{R}^{d}$, with probability measures $\tau$ and $\nu$, respectively, satisfying the equal mass condition $\tau(\Omega) = \nu(\Omega^{*})$. The optimal transport map $T: \Omega \to \Omega^{*}$ is a measure-preserving mapping with the minimization of the total transportation cost,  
\begin{equation}
	\min \int_{\Omega} c(x,T(x))d\tau(x), 
s.t., T_{\#}\tau=\nu , 
\end{equation}
where $ c : \Omega \times \Omega^{*} \rightarrow \mathbb{R}^{+}$ is the cost function, and $_{\#}$ denotes the push forward operator. 

Assume the density functions $f,g$ for $\Omega,\Omega^*$ are given by $d\tau = f(x)dx$ and $d\nu = g(x)dx$, respectively. The OT map is solved by Brenier \cite{brenier1991polar} through the Monge-Ampere equation, 
\begin{equation}
	\det D^{2}u(x) = \frac{f(x)}{g\circ u(x)}, \text{ s.t. } \nabla u(\Omega) = \Omega^{*},
\end{equation}
where the operator $\det D^{2}u$ represents the determinant of the Hessian matrix of the Brenier potential function $u:\Omega \rightarrow \mathbb{R}$, $u \in C^{2}$ (twice continuously differentiable). It was stated that given the cost function $c(x,y) = \frac{1}{2} \left|  x-y \right|^{2} $, the OT map exists and is unique, given by the gradient of $u$, $T = \nabla u$.

Gu et al. \cite{gu2013variational} proposed variational principles for solving discrete Monge-Ampere equations, and gave a general geometric variational approach to the semi-discrete OT problem.

\begin{theorem}[Gu et al.  \cite{gu2013variational}] Let $\Omega$ be a compact convex domain in $\mathbb{R}^{d}$, $\left\lbrace p_{1},p_{2}, \dots ,p_{n} \right\rbrace $ be a set of $n$ distinct points in $\mathbb{R}^{d}$, and $f:\Omega \longrightarrow \mathbb{R}$ be a continuous density function. For any discrete probability measures on $n$ points, $\nu_{1},\nu_{2}, \dots ,\nu_{n}>0$ with $\sum_{i=1}^{n}\nu_{i}=\int_{\Omega}f(x)dx$, there exists a height vector $\textbf{h} = \left( h_{1},h_{2}, \dots ,h_{n}   \right) \in \mathbb{R}^{n}$, unique up to adding  a constant $(c, c, \dots , c)$, such that $\forall \ i \in \left\lbrace  1,\dots ,n\right\rbrace$, 
\begin{equation}\label{measure}
	\tau( W_{i}(\textbf{h})\cap\Omega) = \omega _{i} (\textbf{h}) := \int_{W_{i}(\textbf{h})\cap\Omega}f(x)dx=\nu_{i},
\end{equation}
where the $\tau$-volume $\tau(W_{i}(\textbf{h}))$ denotes the probability measure of each power cell $W_i$. The height vector $\textbf{h}$ is exactly the optimal solution of the convex energy function, 
\begin{equation}
	E(\textbf{h}) = \int_{0}^{\textbf{h}} \sum_{i=1}^{n}\tau( W_{i}(\textbf{h})\cap\Omega)dh_{i}-\sum_{i=1}^{n}h_{i}\nu_{i},
\end{equation}
on the open convex set (the admissible solution space) 
$H = \left\lbrace \textbf{h}\in \mathbb{R}^{n} | \tau( W_{i}(\textbf{h})\cap\Omega) > 0 , \sum_{i=1}^{n}h_{i}=0 \right\rbrace$.
\end{theorem}

The discrete OT map is then given by $T(x)=\nabla u_{\textbf{h}}(x)$. The 
power cell mass center for each power cell $W_{i}$ is calculated as $m_{i}=\int_{\Omega}xd\tau(x)/\nu_{i}$, then 
there is a map $\hat{T}$ induced by the OT map from $\tau$ to $\nu$ : $\hat{T}(m_{i}) = p_{i}, \forall \ i =\left\lbrace  1,\dots ,n\right\rbrace$. 
Detailed calculations can be found in \cite{gu2013variational,su2015optimal,an2019ae}. 

Based on this framework, an extended version \cite{an2019ae} was proposed by specifying zero probability measure to the marginal areas of the distribution, which results in a continuous distribution. We employ the OT merging operation in this work to initialize the distribution of the latent space.  

\subsection{Geometric Mapping}
\label{sec:background-HM}
We introduce the fundamental geometric mappings involved in this work such as Tutte embedding \cite{tutte1963draw} and harmonic mapping \cite{jost2006lectures}, especially focusing on discrete cases.  

Given a connected graph $G=(V_G,E_G)$, assume $G$ has an outer face $o$ which is a simple cycle. Let $V_o \subset V_G$ be the set of vertices on the outer face. \textit{Tutte embedding} of graph $G$ is determined by the boundary condition and the interior barycenter principle. First, embed the vertices of the outer face $V_o$ onto a convex polygon in the plane. Assume the vertices $v_1,v_2,\cdots,v_k$ of $V_o$ are placed at the points $p(v_1),p(v_2),\cdots,p(v_k)$, forming a convex polygon. Then, construct the position of interior vertex $v \in  V_G \setminus  V_o$ as barycenter of its one-ring neighbors $N(v)$, 
\begin{equation}
    p(v) = \frac{1}{|N(v)|} \sum_{u\in N(v)} p(u). 
    \label{eqn:tutte}
\end{equation}
Tutte embedding is equivalent to solving the above linear system, and has theoretical guarantee of existence and uniqueness.

Assume a triangular mesh is denoted as \( M=(V,E,F) \), where \( V, E, F \) represent the vertex, edge and triangular face sets of the mesh, respectively. \textit{Harmonic mapping} is defined as \( h:V\longrightarrow \mathbb{\mathbb{R}}^2 \), and is achieved when the following harmonic energy is minimized:
\begin{equation}
	E(h) = \frac{1}{2} \sum_{ [v_{i},v_{j}] \in E} w_{ij}(h(v_{i})-h(v_{j}))^{2}.
\end{equation}
Here, $w_{ij}$ is the cotangent edge weight given by
\begin{equation} \label{edge weight}
	w_{ij}=\left\{
	\begin{aligned}
		&cot\theta_{ij}^{k} + cot\theta_{ij}^{l} , &e_{ij} \notin \partial M \\
		&cot\theta_{ij}^{k}, &e_{ij} \in \partial M
	\end{aligned},
	\right.
\end{equation}
where the \( \partial M \) indicates the boundary of the triangular mesh, and \( \theta_{ij}^{k} \) represents the corner angle in face \( [ v_{i},v_{j},v_{k} ] \) at \( v_{k} \). 
Then, the discrete Laplace operator $\bigtriangleup f$ is defined as follows:
\begin{equation}
	\bigtriangleup h(v_{i}) = \sum_{ [v_{i},v_{j}] \in E} w_{ij}(h(v_{i})-h(v_{j})),
      \label{eqn:harmonic_fundtion}
\end{equation}
and $h$ is obtained by solving the sparse linear system. When \( \Delta h = 0 \), \( E(h) \) reaches a minimum value.
For a topological disk triangular mesh, when the target domain is convex, its harmonic mapping exists and is unique, and is guaranteed to be a diffeomorphism. 

It can be seen that both the Tutte embedding of graphs and the harmonic mapping of triangular meshes are ultimately solved by a linear system, the main difference between the two being the setting of the edge weights, which are the vertex degree in the former and the cotangent angle in the latter. When dealing with a surface decorated with a feature graph, the harmonic mapping can be extended by combining the Tutte embedding strategy on the feature graph to obtain a canonical convex-subdivision constrained harmonic mapping  \cite{Zeng2015Graph,Zeng2018Graph}.

%% file: 4_algorithm.tex
\section{Algorithms}
\label{sec:algorithm}
\begin{figure}[t]
	\centering
	\includegraphics[width=0.7\linewidth]{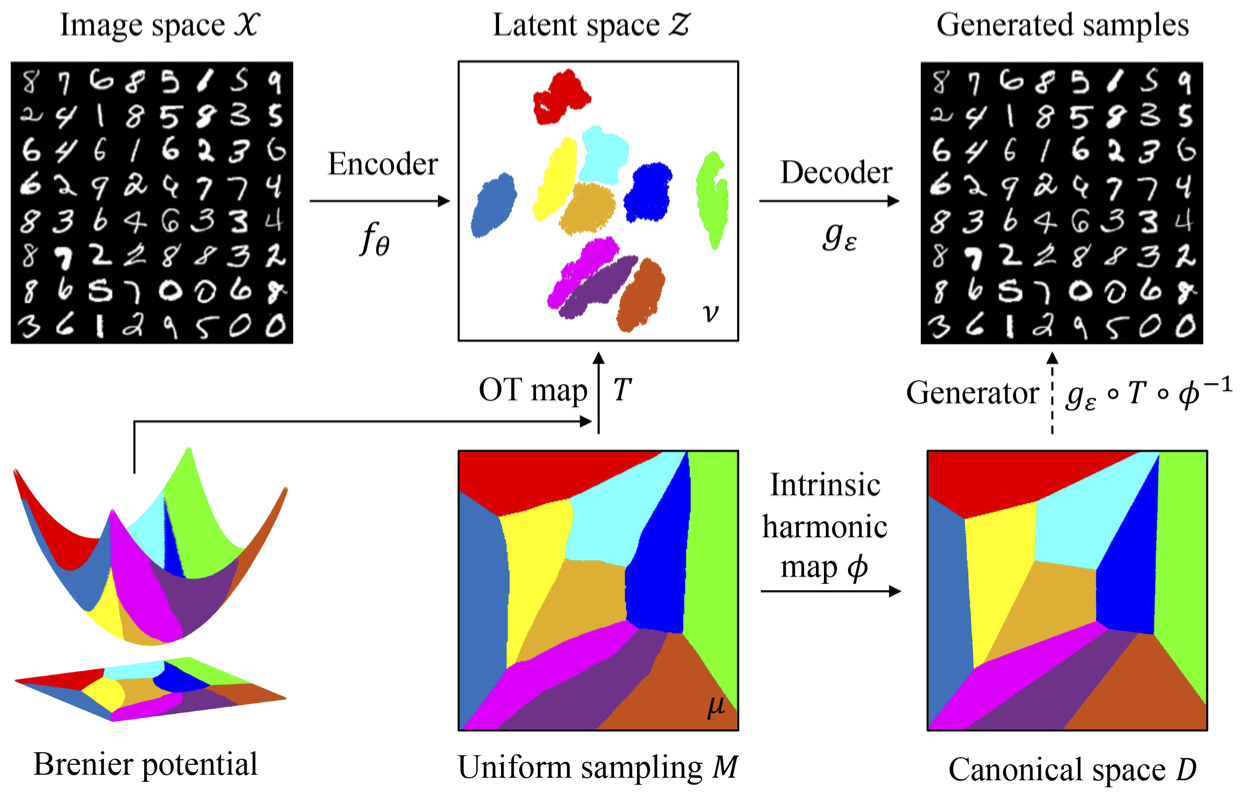}
	\caption{Workflow of building canonical structural representation of latent space and generative model for a single domain (one dataset). }
        \label{fig:single}
\end{figure}

The proposed cross-domain generative model is carried out by incorporating geometric transformations into latent spaces within the deep encoder-decoder architecture (see Fig. \ref{fig:cross}). 
The canonical representation of latent space based on optimal transport and harmonic map can also be integrated into the encoder-decoder pipeline to accomplish single-domain generation (see Fig. \ref{fig:single}). 
The whole generative model comprises three major modules:

\vspace{1mm}
\textit{AutoEncoder.} An autoencoder is trained to encode the data manifold, denoted as $f_{\theta}$, from the image space $\mathcal{X}$ to the latent space $\mathcal{Z}$. The decoder $g_{\varepsilon}$ then decodes the latent code back to the data manifold (see Fig. \ref{fig:AE}).

\vspace{1mm}
\textit{Canonical representation of latent space.} 
For a single latent space, we convert it to a canonical representation by performing the following operations (see Fig. \ref{fig:Mapper}): 
\begin{enumerate}
    \item Perform barycentric translations $t$ to separate latent clusters without overlapping; 
    \item Compute optimal transport $o$ to merge latent clusters while eliminating in-between margins, so that the converted latent space has connected clusters, whose boundaries form a feature graph of the space; and
    \item Compute graph-constrained harmonic mapping $\phi$ to transform the merged space to a convex-subdivision parameter domain (named canonical latent space), where the curvy feature graph becomes a planar straight-line graph with convex faces.
\end{enumerate}

This procedure maps triangular mesh $M$ constructed from optimal transport merging result (with uniform distribution) onto the canonical convex subdivision domain $D$ 
by minimizing harmonic energy with graph constraints.

\vspace{1mm}
\textit{Registration between latent spaces with cluster constraints.}  For different latent spaces ($L_i,i=1,2$) with given latent cluster correspondences, we achieve their bijective alignment by the following strategy (see Fig. \ref{fig:cross}): 
\begin{enumerate}
    \item Conduct barycentric translations $t_i$ to make the latent spaces have consistent topologies, i.e., the same neighboring relationship of clusters;
    \item Compute canonical convex-subdivision representations by $\phi_i$ (with merging $o_i$) on their translated latent spaces;
    \item Compute convex-subdivision constrained harmonic registration $h$ between their canonical representations; and 
    \item Establish the final alignment $f$ from source to target through the composition of all the above transformations, given by $f:=t_2^{-1} \circ o_2^{-1} \circ \phi_2^{-1} \circ h \circ \phi_1 \circ o_1 \circ t_1$.
\end{enumerate}

\subsection{Canonical representation of a single latent space}
\paragraph{Barycentric translation}
For a single latent space, if clusters have overlaps, we differentiate them into isolated clusters simply by a set of translations. We first compute the barycenter $c_i$ and the largest radius $r_i$ for each cluster $C_i$, and then compare the distance between barycenters $d_{ij}=|c_i-c_j|$ and the summation of their radii $r_i,r_j$, $r_{ij}=r_i+r_j$ for every cluster pair. The \textit{cluster separation condition} is defined as:  
\begin{equation}
d_{ij} \geq r_{ij}.
\label{eqn:sep_condition}
\end{equation}

We repeatedly traverse all the clusters in order until all cluster pairs satisfy the separation condition: for cluster $C_i$, check all other clusters $C_j$. If $d_{ij}<r_{ij}$, then move cluster $C_j$ farther from cluster $C_i$ by $r_{ij}-d_{ij}$ along the center line.

\vspace{1mm}
\paragraph{Optimal transport merging} 
Based on the description in Section \ref{sec:background-OT}, the key of OT is to compute the $\tau$-volume $w_i(\textbf{h})$ of each cell $W_i(\textbf{h})$, which can be estimated using conventional uniform grid sampling method. Draw $N$ samples from $\tau$ distribution within $W_i$, then the estimated $\tau$-volume is $\hat{w}_i(\textbf{h}) = (\sum _j x_j)/N, x_j\in W_i(\textbf{h})$. 
When $N$ is large enough, $\hat{w}_i(\textbf{h})$ converges to $w_i(\textbf{h})$. Then the gradient of the energy is approximated as $\nabla E(\textbf{h}) \approx (\hat{w}_i(\textbf{h})-\nu_i)^T$. Once the gradient is estimated, Adam algorithm is used to minimize the energy. In practice, if the energy $E(\textbf{h})$ stops decreasing for a number of consecutive steps, then interpolate sample points. Intuitively, the final result demonstrates the fact that the margins are merged and the samples are uniformly distributed. More computational details of semi-discrete OT map and its extended version can refer to \cite{gu2013variational,an2019ae}, respectively.

\vspace{1mm}
\paragraph{Graph-constrained harmonic mapping}
We construct the Delaunay triangulation on the sample points obtained in OT merging step to have the triangular mesh $M=(V,E,F)$, where the cluster boundaries are labeled, forming a graph $G=(V_{ G}, E_{G},F_{G})$. Then we map the graph-decorated surface $(M, G)$ onto the convex-subdivision domain $D$ by minimizing harmonic energy with the graph constraints, $\phi:(M,G) \rightarrow (D,\hat{G})$, where $G$ is intrinsically converted to a convex-subdivision $\hat{G}=(V_{\hat G}, E_{\hat G},F_{\hat G})$. 

As described in Section \ref{sec:background-HM}, the critical point of harmonic energy is harmonic map. The energy is formulated as Eqn. (\ref{eqn:harmonic_fundtion}), where the setting of cotangent edge weight implies the harmonic property. 
For the graph-constrained problem, we employ special handling on neighbors and edge weights to automatically and intrinsically map the curvy graph $G$ as a convex subdivision $\hat{G}$ on the unit square. 

{\setlength{\tabcolsep}{1pt}
\begin{figure}[htbp]
\footnotesize
    \centering
    \begin{tabular}{cc} 
\includegraphics[height =0.15\textwidth]{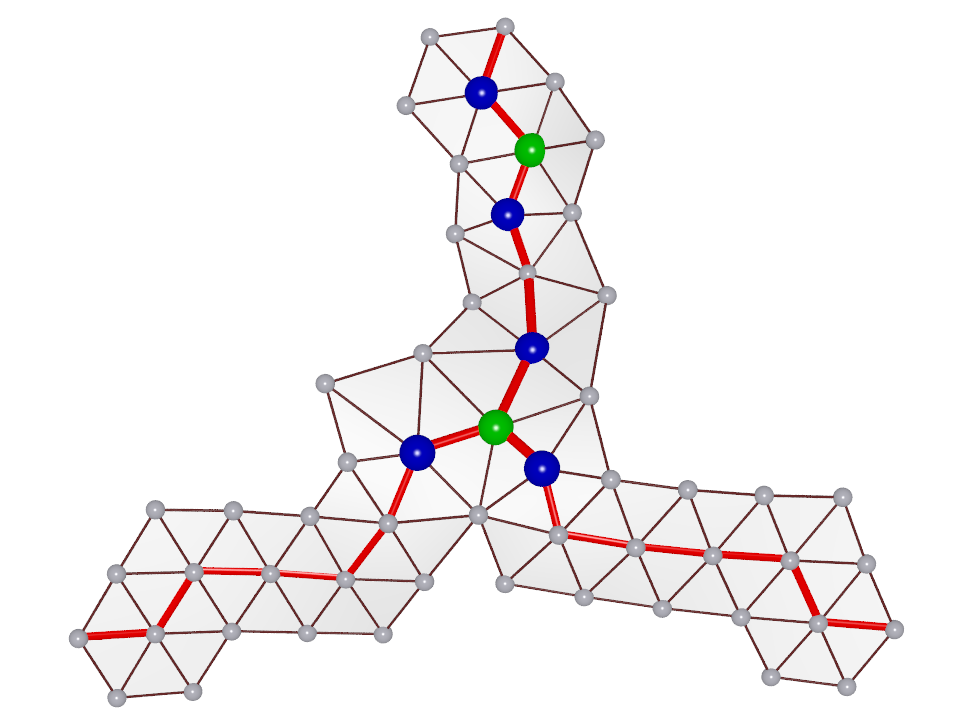} & 
\includegraphics[height=0.15\textwidth]{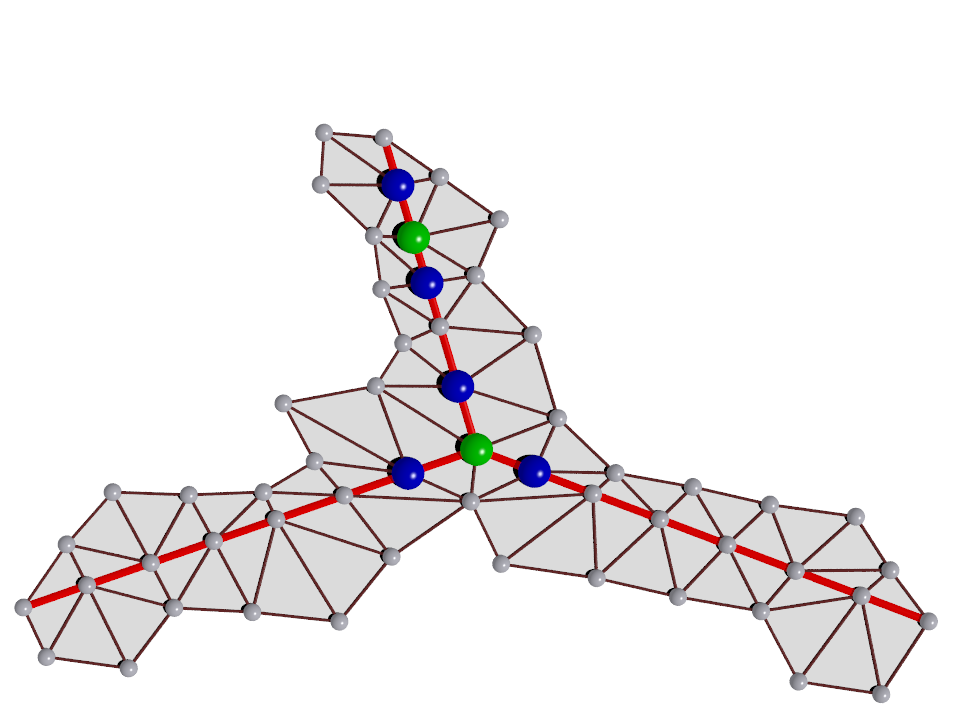} \\ 
       (a) Original mesh & (b) Straightened mesh \\
      \end{tabular}
    \caption{Graph-constrained harmonic mapping (local mesh structures).}
    \label{fig:graph-edge-weight}
\end{figure}
}

If the vertex $v_0$ in the triangular mesh (see Fig. \ref{fig:graph-edge-weight}) is: 
\begin{enumerate}
\item not on the graph (i.e.,  unlabeled interior vertex in the mesh), then we use the cotangent edge weight 
on its one-ring neighboring vertices on the mesh; or
\item lying inside the interior of graph edge, then we compute the barycentric coordinate (denoting edge length ratio) by its one-ring graph neighborhood (i.e., the previous and next neighbors on the graph edge); or 
\item the graph node, then we apply the circumferential mean value theorem \cite{floater2003mean} to its one-ring graph neighborhood  (i.e., the neighboring vertices on the graph) to compute the adaptive harmonic weight.
\end{enumerate}

With the above setting, the constrained harmonic map is computed in one-shot by solving a sparse linear system. The resulted map exists and is unique, and it is diffeomorphic and intrinsic, respecting the original geometry. The algorithm is easy to be implemented and the computation is quite efficient. 
More computational details can be found in \cite{Zeng2015Graph,Zeng2018Graph}.

\subsection{Alignment between two latent spaces}
\label{method:translation}

\paragraph{Barycentric translation} 
We design transform latent clusters through barycentric translations to ensure the OT merging results of source and target latent spaces have a consistent topology of clusters. This is accomplished in two ways: i) by transforming one domain based on the other; ii) by transforming both domains to a specified layout.

\vspace{1mm}
\textit{Transform one domain.} We first remove the outliers and overlaps of latent clusters for both latent spaces, to make sure all clusters are separate, then fix target domain and compute its OT merging result. We translate source clusters corresponding to the layout of target domain. 
Iterative translations may be required, described as follows: 
\begin{enumerate}
    \item Move the source clusters to the corresponding target positions and adjust the distance between clusters to satisfy the \textit{cluster separation condition} $d_{ij}\geq r_{ij}$;
    \item Calculate the OT merging of the translated source domain and check for consistency with the target OT result: If yes, then return the current source layout; Otherwise, adjust the inconsistent clusters on source based on the position relationship. Concretely, suppose that the source clusters A and B are not adjacent to each other in OT merging as that in target layout.  
    We can slightly translate cluster A along the center line towards cluster B and keep the \textit{cluster separation condition} satisfied at the same time to make them adjacent in OT merging. 
\end{enumerate}

\begin{figure}[!ht]
	\centering
	\includegraphics[height=0.2\linewidth]{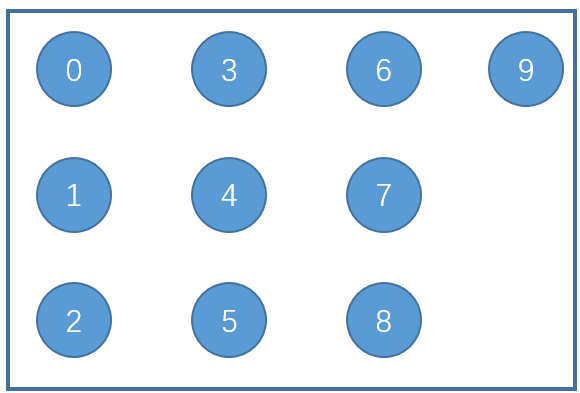}
	\caption{Transform both domains to the same layout.} 
        \label{fig:translation_2}
\end{figure}

\vspace{1mm}
\textit{Transform both domains.} We expect the two domains to have similar layout, namely the corresponding clusters have the same relative position relationship. To achieve this, we can translate clusters into a specified configuration, here we choose the regular grid layout as shown in Fig. \ref{fig:translation_2} (assume there are 10 clusters). 
For each domain, let $R$ be the largest cluster radius among all clusters, $R = \max r_i$, we set the distance between horizontally or vertically adjacent clusters be $d=2R$. This makes that the clusters in regular layout are separated.  
It is experimentally shown that this layout on both domains makes their OT results topologically consistent.

\vspace{1mm}
\paragraph{Graph-constrained harmonic registration}
With the defined consistent feature graphs, the source $(M_1,G_1)$ and the target $(M_2,G_2)$ are mapped onto the square domains with interior convex-subdivisions by the previously mentioned intrinsic graph-constrained harmonic maps.  
Then the graph-constrained harmonic mapping is performed between the two convex-subdivision domains, $h:(D_1,\hat{G}_1) \rightarrow (D_2,\hat{G}_2)$. The final registration is achieved by the compositions of mappings, $f':= \phi_2 ^ {-1}\circ h \circ \phi_1$, as in the following diagram:
\[
\begin{CD}
(M_1,G_1) @>f'>>  (M_2,G_2) \\
@V{\phi_1}VV @VV{\phi_2}V\\
(D_1,\hat G_1) @>h>> (D_2, \hat G_2)
\end{CD}
\label{eqn:diagram_0}
\]

In the minimization of harmonic energy with convex-subdivision constraints, we similarly integrate the constraints into the setting of edges weight to ensure that the vertices on graph edge slide smoothly on the corresponding target graph edge. We specify the positions of the boundary vertices $v_s \in \partial D_1$ (by interpolation on the corresponding target edge) and the graph nodes $v_s \in V_{\hat G_1}$ (as the corresponding target ones), and set edge weight for interior vertices on graph edge $e_{\hat G_1}\in E_{\hat G_1}$ only using adjacent edges on graph. For other interior vertices, we use cotangent edge weights. In detail, the harmonic function $h$ is computed as follows: 
\[
\bigtriangleup  h(v_i)=\sum_{[v_i,v_j]\in E_1} w_{ij}(h(v_i)-h(v_j))=0, v_i \in V_1,
\]\[
h(v_s)=v_t, v_s\in V_{\hat G_1}, v_t\in V_{\hat G_2},
\]
\[
h(\partial_{D_1})=h(\partial_{D_2}), h(e_{\hat G_1})=h(e_{\hat G_2}), e_{\hat G_k}\in E_{\hat G_k}.
\]
The resulted mapping has the same properties as the canonical graph-constrained harmonic map, namely, the map exists, and is unique and diffeomorphic \cite{Zeng2018Graph}.

%% file: 5_model.tex
\section{Models}
\label{sec:models}

This section introduces typical generative models based on the GMapLatent framework presented in Section \ref{sec:algorithm} in the context of latent space clustering distributions. Here, we focus on the design of latent space alignment, which is the core of the proposed encoder-decoder based cross-domain generation architecture. We illustrate the models on the binary image datasets of handwritten digits, by translating Chinese MNIST (source domain) to Arabic MNIST (target domain). As shown in Fig. \ref{fig:latent_space}, latent codes of a class do not always come together to form a tight cluster, and clusters usually have outliers and irregular shapes. Whether or not outliers are handled affects the performance of generation. 

\begin{figure}[t]
    \centering
    \includegraphics[width=0.7\linewidth]{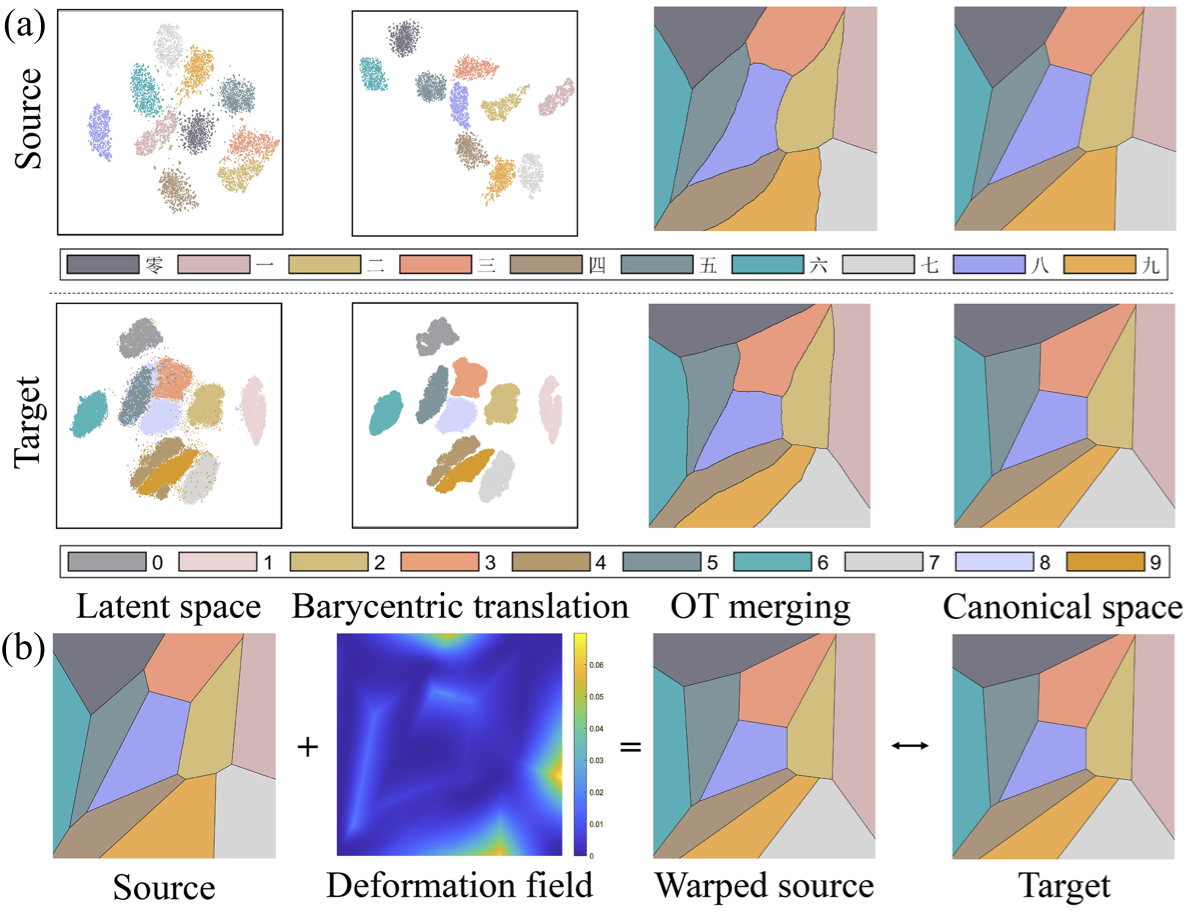}
    \caption{Workflow of outlier-free GMapLatent: without outliers.}
    \label{fig:o_1}
\end{figure}

\begin{figure}[t]
    \centering
    \includegraphics[width=0.7\linewidth]{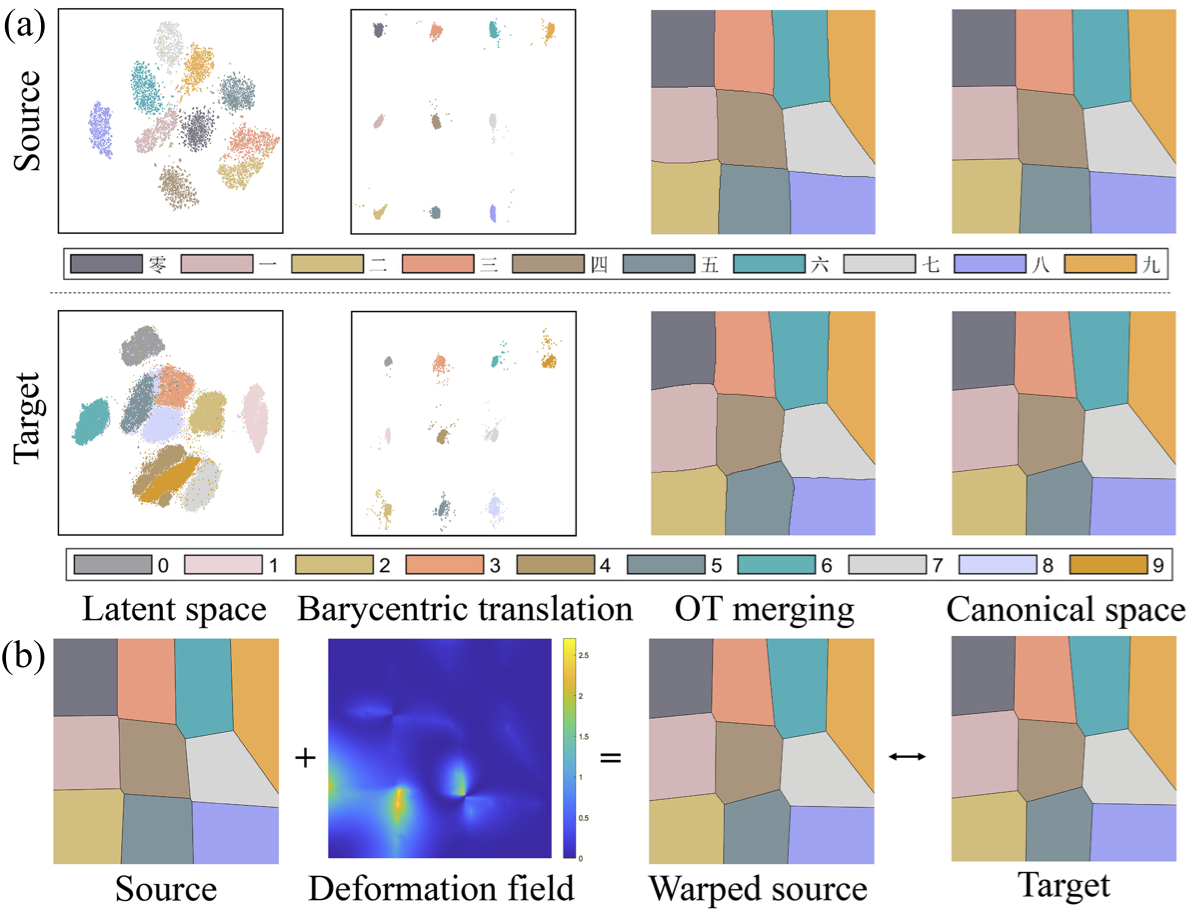}
    \caption{Workflow of full-domain GMapLatent: with outliers.}
    \label{fig:o_2}
\end{figure}

\subsection{Outlier-Free GMapLatent: without outliers} 
A straightforward design approach is to simplify the problem by removing the outliers shown in the 2D latent space.  
Figure \ref{fig:o_1} depicts the workflow of the outlier-free model. We first remove the outliers from both domains, retaining the main aggregated portion of each class, such that all clusters are separate. We then fix the target latent space and perform barycenter translations on the source latent space using the method in Section \ref{method:translation}, in preparation for the required graph isomorphism in the alignment. 
With the proposed strategy, the OT merging results indicate that the adjacency relationships between the corresponding clusters remain consistent. Subsequently, 
the straightened source is registered to the straightened target.  
Through the diffeomorphic correspondences and the transformations performed, the alignment between the original latent spaces is achieved.

\subsection{Full-Domain GMapLatent: with outliers} 
Assume that all samples are useful data. Outliers shown in 2D may be caused by dimensionality reduction techniques or from unbalanced sampling. 
In order to be able to fully utilize all the given data information and to demonstrate the practical applicability of our model, we consider full-domain latent codes in our design, taking into account all samples in the training dataset. The key difference from the outlier-free model lies in the preprocessing of latent space. We adapt both source and target latent spaces by barycentric translations without outlier cleaning using the strategy in Section \ref{method:translation}. 
Figure \ref{fig:o_2} depicts the workflow of the model with outliers. The barycentric translation results show that the clusters are moved to the corresponding centers of uniform grids to ensure that the clusters are separate and the clusters have the same spatial layout. The OT merging results indicate that the feature graphs are isomorphic, where both the intra-cluster and inter-cluster margins are merged, intuitively. 
The subsequent steps and the alignment computation are similar to the outlier-free model. 

%% file: 6_experiment.tex
 \section{Experimental results}
 \label{sec:experiment}
We conducted experiments on GMapLatent model without or with outliers and compared their performance with existing GAN-based and OT-based models. 

Two evaluation metrics are employed: (1) \textit{FID} (Frechet Inception Distance) \cite{yu2021frechet}, which measures the difference between the distribution of the generated image and that of the real image. The smaller the FID value, the closer the generated image is to the distribution of the real images; and (2) \textit{Accuracy}, which reflects the probability that the source domain image is correctly translated into the real category of the target domain. In the computation process, we train a LeNet classifier \cite{el2017cnn} on the real data of the target domain to predict the category labels of the translated images, and calculate the accuracy of the prediction based on the corresponding translated labels.

\subsection{Binary handwritten images}
We performed a detailed experimental demonstration using the translation task of binary handwritten digit images, from the Chinese MNIST \cite{ChineseMnist}  data domain to the Arabic MNIST \cite{deng2012mnist} data domain. These two datasets have specified training and test sets.  
In the network architecture, the encoder includes five convolution layers for Chinese MNIST and Arabic MNIST, as shown in Table \ref{c_en}. It adopts the mirrored architecture of the decoder, which has the same architecture as the consistent generator in GAN. 
The batch size, the epoch, and the learning rate in Chinese and Arabic MNIST autoencoders are set to (150, 500, 2e-4) and (120, 500, 9e-5), respectively.

\begin{table}[!ht]
\centering
    \caption{Encoder architecture of autoencoder on MNIST datasets.}
    \label{c_en}
{
    \begin{tblr}{
  hline{1,8,14} = {-}{0.08em},
  hline{2} = {-}{0.05em},
}
Layer    & \#Outputs & \#Kernel  & Stride & BN  & Activation \\
Input (Chinese) & 64*64*1           &             &        &     &            \\
Convolution          & 32*32*$d$     & 4*4         & 2      &     & LeakyReLU  \\
Convolution          & 16*16*$d$*2   & 4*4         & 2      & Yes & LeakyReLU  \\
Convolution          & 8*8*$d$*4     & 4*4         & 2      & Yes & LeakyReLU  \\
Convolution          & 4*4*$d$*8     & 4*4         & 2      & Yes & LeakyReLU  \\
Convolution          & 150               & 4*4         & 1      &     &   \\   
Input (Arabic)  & 28*28*1 & & & & \\
    Convolution & 14*14*$d$ & 4*4 & 2 & & LeakyReLU \\ 
    Convolution &7*7*$d$*2 &4*4 &2 &Yes &LeakyReLU \\
    Convolution &4*4*$d$*4 &4*4 &2 &Yes &LeakyReLU \\
    Convolution &2*2*$d$*8 &4*4 &2 &Yes &LeakyReLU \\
    Convolution &150 &2*2 &1 & &      \\
\end{tblr}
}
\end{table}

\vspace{1mm}
\paragraph{Point-to-point generation} 
Figure \ref{res_img} shows the translation images of cycle-GAN and outlier-free GMapLatent (without outliers) and full-domain GMapLatent (with outliers) for a set of test samples. Although the images translated by the cycle-GAN are clear, the vast majority of the test source domain images are incorrectly translated into other figures. Our method can translate with higher accuracy and quality. Table \ref{tab:com} demonstrates that numerically. It can be seen that full-domain GMapLatent (with outliers) performs better than outlier-free GMapLatent (without outliers), and it achieves the highest accuracy and comparable FID among existing methods. 
Comparing to other existing best performance, the accuracy is significantly improved by 18.36 percentage points. These show that GMapLatent can better convert the source image to the corresponding target image. 

{\setlength{\tabcolsep}{1pt}
\begin{figure}[t]
\footnotesize
    \centering
    \footnotesize
\begin{tabular}{cc}
        \centering
        \includegraphics[width=0.48\linewidth]{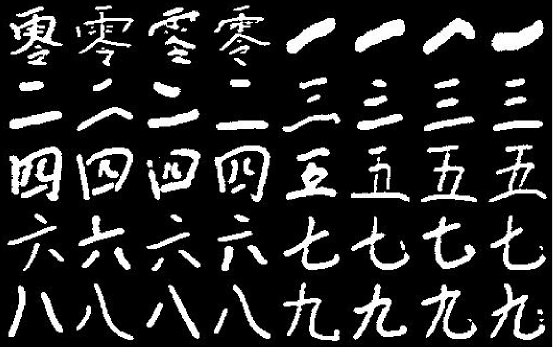}&
        \includegraphics[width=0.48\linewidth]{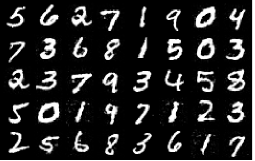}\\
 Input test data & Output of cycle-GAN \\        
 \includegraphics[width=0.48\linewidth]{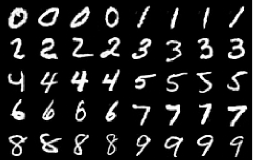} &  
 \includegraphics[width=0.48\linewidth]{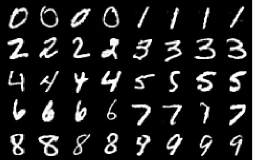}\\
  Output of GMapLatent (w/o outliers) &  Output of GMapLatent (w outliers)\\
 \end{tabular}
\caption{Translation images of cycle-GAN and outlier-free GMapLatent (w/o outliers) and full-domain GMapLatent (w outliers).}
    \label{res_img}
\end{figure}
}

\begin{table}[!ht]
\centering
    \caption{Comparison with other methods: \textit{Acc} - \textit{Accuracy} ($\%$). }
    \label{tab:com}
    {
    \begin{tblr}{
      hline{1,2,11} = {-}{0.08em},
      hline{3} = {-}{0.05em},
    }
    &Digit &&Animal&\\
    Method &\textit{FID} $\downarrow$ &\textit{Acc} $\uparrow$&\textit{FID} $\downarrow$ &\textit{Acc} $\uparrow$ \\
    Cycle-GAN \cite{zhu2017unpaired}  &6.99  &22.72 &78.56  &30.27\\
    TCR \cite{mustafa2020transformation}  &6.90  &36.21&342.48  &33.33 \\
    W2GAN \cite{korotin2019wasserstein}  &12.04  &34.21 &121.86  &28.40  \\
    OT-ICNN \cite{makkuva2020optimal}  &14.37  &29.12 &126.43  &34.67 \\
    KPG-RL-BP \cite{gu2022keypoint}  &157.38  &74.51 &285.43  &60.00 \\
    KPG-RL-MBP \cite{gu2022keypoint} &\textbf{6.54}  &76.14 &\underline{81.02}  &77.27\\
    \textbf{GMapLatent (\textit{outlier-free})}  &83.73  &\underline{94.20}  &101.12 &\underline{85.79}\\
    \textbf{GMapLatent (\textit{full-domain})}  &\underline{13.75} &\textbf{94.50} &\textbf{76.79}  &\textbf{86.15}\\
    \end{tblr}
}
\end{table}

{\setlength{\tabcolsep}{1pt}
\begin{figure}[htbp]
\footnotesize
    \centering
    \begin{tabular}{ccc} 
        \includegraphics[height =0.15\textwidth]{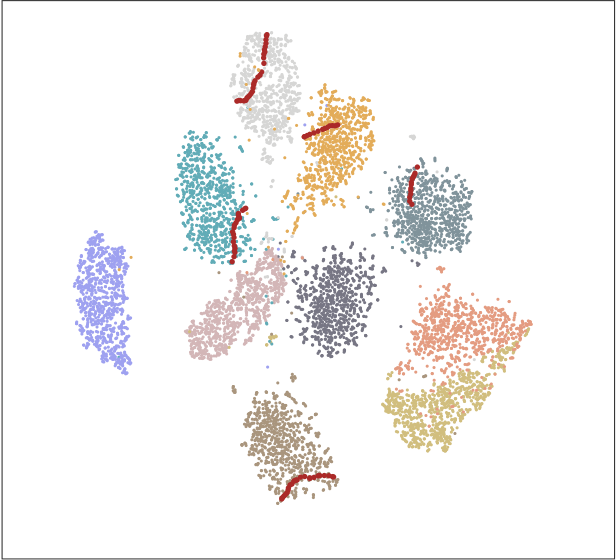} & 
        \includegraphics[height=0.15\textwidth]{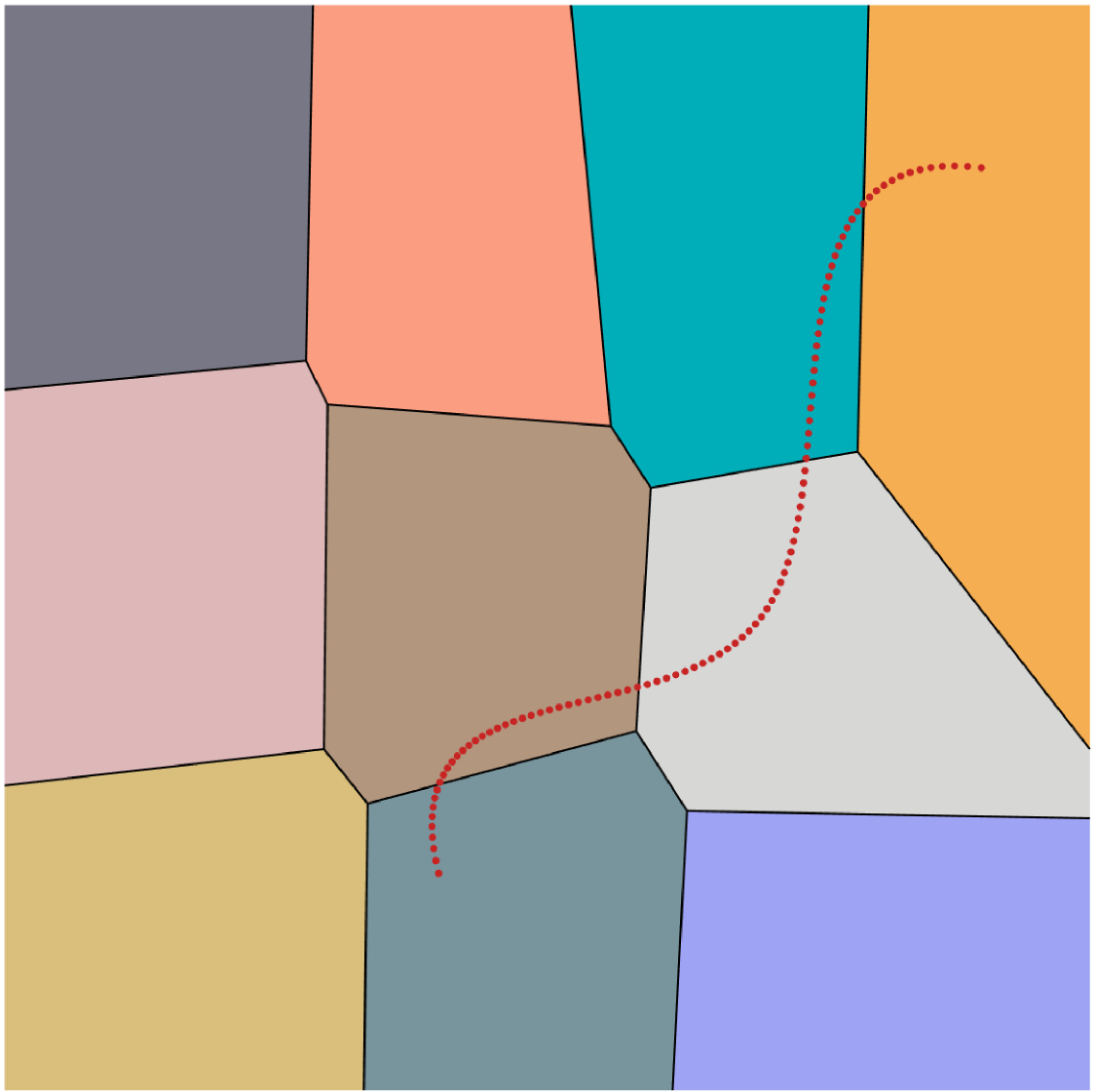} & 
        \includegraphics[height=0.15\textwidth]{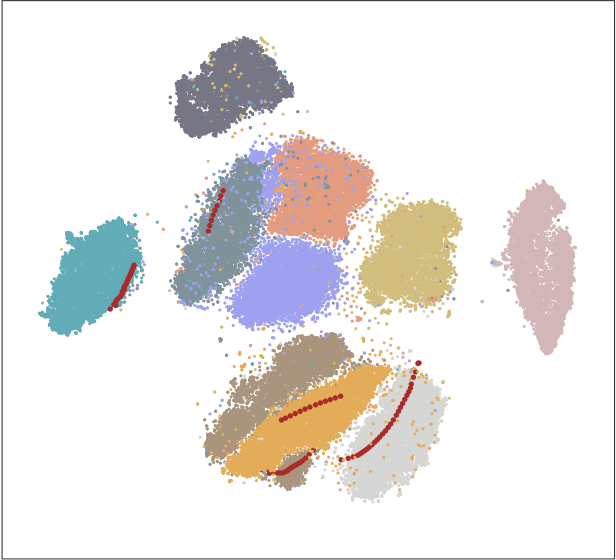} \\
      Source & Registered space & Target \\
    \end{tabular}
    \begin{tabular}{cc}
        \includegraphics[width=0.49\textwidth]{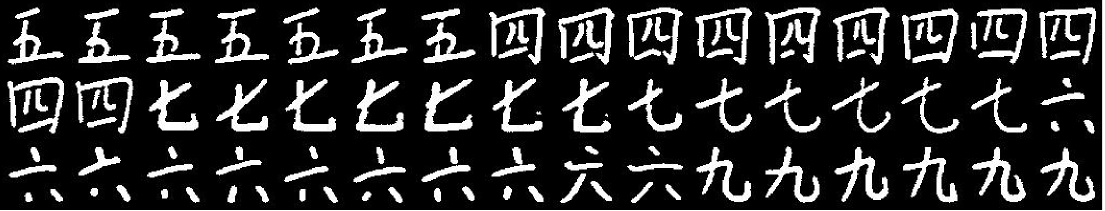}\\
        \includegraphics[width=0.49\textwidth]{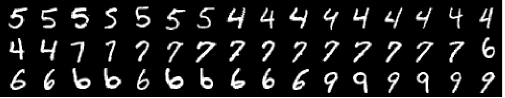}\\
        Translation images along the path
    \end{tabular}
    \caption{Curve-to-curve generation through full-domain GMapLatent.}
    \label{fig:curve-to-curve}
\end{figure}
}

\vspace{1mm}
\paragraph{Curve-to-curve generation} 
Besides point-to-point generation, we validate the necessity of the proposed alignment by considering the case of curve-to-curve generation, because it requires translating all  samples on the curve. Without OT merging, the samples on a curve drawn in the latent space may pass through the margins of clusters, leading to model collapse. By OT merging, a curve is smoothly translated to a sequence of semantically meaningful images across classes or within a class. Figure \ref{fig:curve-to-curve} gives the translation sequences and the curve distribution in the original source and target latent spaces, demonstrating the accurate alignment of source and target categories. Therefore, this canonical representation and precise registration in common registered space make curve interactions feasible in a way  other methods cannot. 

\subsection{Natural color images} 
\begin{figure*}[t]
    \centering
    \includegraphics[width=0.8\linewidth]{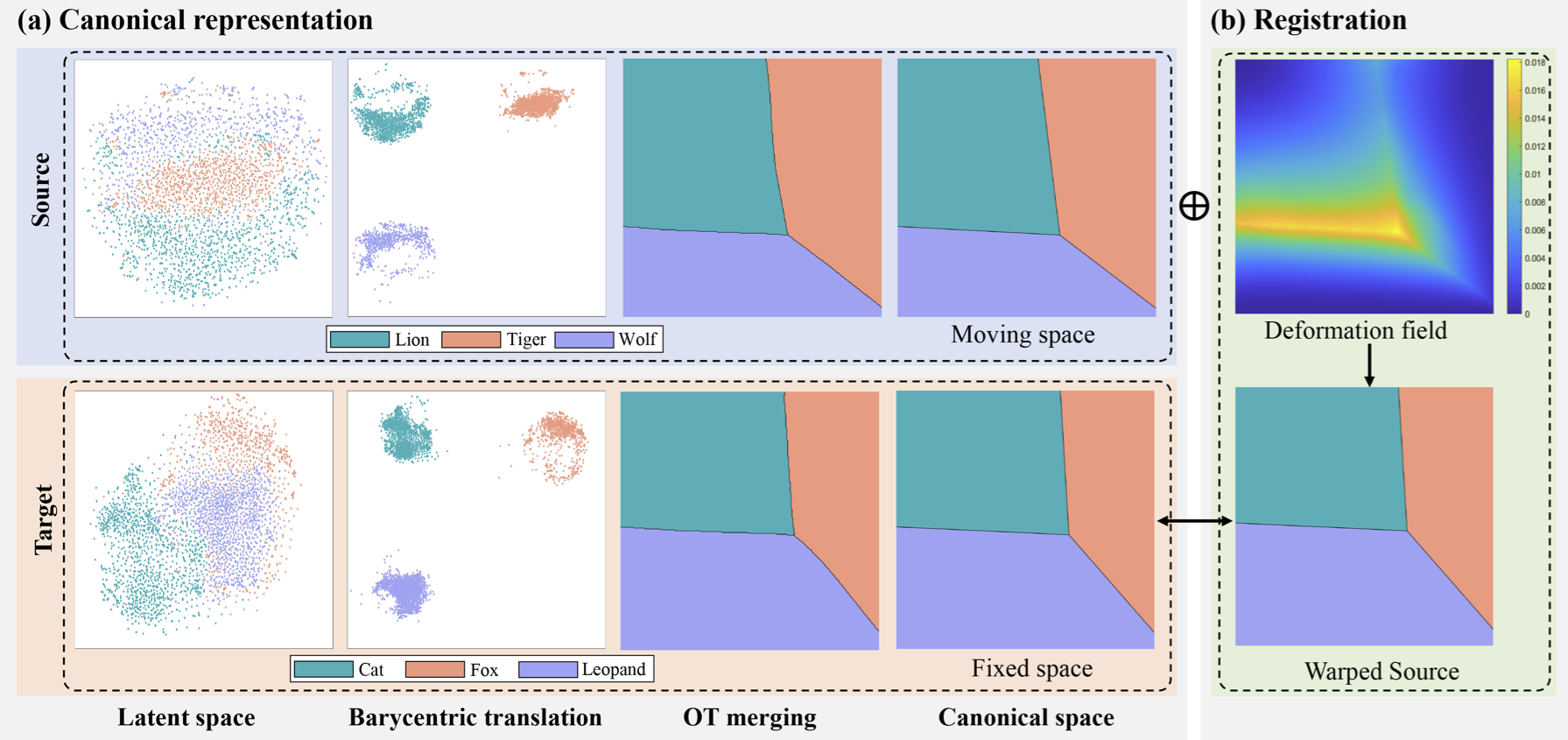}
    \caption{Full-domain GMapLatent for color images. Source: (Lion, Tiger, Wolf); Target: (Cat, Fox, Leopard).}
    \label{fig:afhq_1-1}
\end{figure*}

We also tried to validate our model on color images with more complicated contents. We employed Animal Faces-HQ (AFHQ) dataset \cite{choi2020stargan}, and selected three species (Lion, Tiger, Wolf) as source domain and three other species (Cat, Fox, Leopard) as target domain, where 1000 images were randomly collected for each specie.
In the network architecture, the decoder has the same as the consistent generator architectures in bigGAN \cite{brock2018large}, and the encoder has the mirrored architecture. The full-domain GMapLatent workflow on this setting is shown in Fig. \ref{fig:afhq_1-1}. 
We also analyzed the experimental results qualitatively and quantitatively, shown in Fig. \ref{res_img1} and Table \ref{tab:com}, respectively. 
Similar to the performance on binary handwritten images, the full-domain GMapLatent achieves the highest accuracy, significantly improved by 8.88 percentage points comparing to other exisiting best performance. In addition, it obtains the lowest FID.

{\setlength{\tabcolsep}{1pt}
\begin{figure*}[ht]
    \centering
    \footnotesize
    \begin{tabular}{ccc}
        \centering
        \includegraphics[width=0.33\textwidth]{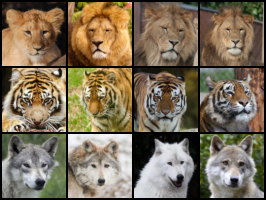}
&
        \includegraphics[width=0.33\textwidth]{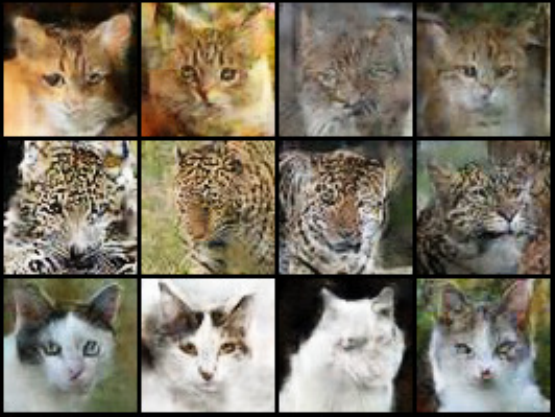}
&
        \includegraphics[width=0.33\textwidth]{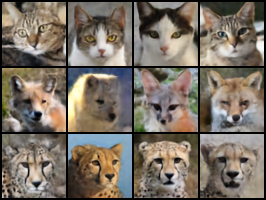}
\\
    Input test data & Output of cycle-GAN & Output of GMapLatent \\        
    \end{tabular}
    \caption{Translation images of cycle-GAN and full-domain GMapLatent.}
    \label{res_img1}
\end{figure*}
}

\subsection{Ablation Studies}
We perform ablation studies on the core modules of GMapLatent, \textit{canonical latent representation} and \textit{precise geometric registration} in our full-domain generative model. 
Here, we start with DirectAlign that performs direct alignment between the transformed latent spaces without the special registration operation. 
Table \ref{tab:com_abla} shows the corresponding numerical results. 
We first introduce the module of canonical latent representation into DirectAlign and the results exhibit better performance in terms of both Accuracy and FID metrics than the irregularly shaped OT maps. 
We further employ the module of precise geometric registration into the model and the results significant improves on the accuracy by 7.4 and 4.2 percentage points on two data cases, respectively, with comparable FID at the same time. 
All the above verifies the efficacy of both modules in the proposed cross-domain generative model. 

\begin{table}[!ht]
\centering
    \caption{Ablation results: canonical representation (CR); graph-constrained harmonic registration  (GCHR).}
    \label{tab:com_abla}
    {
    \begin{tblr}{
  hline{1,2,3,6} = {-}{0.08em},
  hline{3} = {-}{0.05em},
}
    &Digit &&Animal&\\
    \textbf{Method} &\textit{FID} $\downarrow$ &\textit{Acc} $\uparrow$ &\textit{FID} $\downarrow$ &\textit{Acc} $\uparrow$ \\
    DirectAlign (\textit{w/o CR; w/o GCHR})  &15.37  &85.53  &\underline{75.12} &80.29\\
    DirectAlign (\textit{w CR; w/o GCHR})  &\underline{13.81}  &\underline{87.12}  &\textbf{75.10} &\underline{81.94}\\
    \textbf{GMapLatent (\textit{w CR $\&$ GCHR})}  &\textbf{13.75}  &\textbf{94.50} &76.79  &\textbf{86.15}\\
    \end{tblr}
}
\end{table}

\subsection{Discussion} 
\paragraph{Accuracy \& FID} 
Overall, the proposed model has higher accuracy than other state-of-the-art generative models with comparable FIDs. 
Cycle-GAN, TCR, W2GAN, and OT-ICNN models have lower accuracy due to the fact that these methods do not introduce the correlation information about the categories in source and target domains. The  KPG-RL models have relatively higher accuracy because that they use keypoints to guide the category pairing, which improves the translation accuracy.  
Misleading may occur when the keypoints are close to the boundaries of the category clusters and close to other category clusters, and the clusters are not fully aligned. In contrast, GMapLatent can precisely register category clusters including boundaries and interiors, where one-to-one and onto category correspondences are accurately constructed, which is the reason why GMapLatent obtains higher accuracy. It is worth noting that not removing the outliers improves the quality (FID) and accuracy of the generated images. This is because that not removing outliers preserves more patterns (modes), which increases the diversity of the generated samples and thus improves the FID. Also, not removing outliers preserves more information about the cluster boundaries in the latent space, which allows GMapLatent to better predict the classes of test points that are close to the cluster boundaries.

\vspace{1mm}
\paragraph{Sampling strategy \& data augmentation} 
In GMaplatent, canonical latent representation facilitates sampling for controllable generation. 
That is because: 
all latent clusters have been separated by the barycentric translation preprocessing, and are connected only along boundaries of convex polygons. Sampling on such a canonical representation can easily distinguish the class types and avoid mode mixture at this level. Thus, it is feasible to use a given class type or intra/inter-class curve to guide sampling. Therefore, the canonical representation helps provide an auxiliary sampling strategy and make the generation interpretable. It can be used for both cross-domain (see Fig. \ref{fig:cross}) and single-domain (see Fig. \ref{fig:single}) data augmentation with controllable guidance. 

\vspace{1mm}
\paragraph{Difference from AE-OT} 
We recall that AE-OT \cite{an2019ae} was designed for single-domain generation, which performs cluster cleaning and OT merging for generation, and can avoid mode collapse. GMapLatent is designed for cross-domain generation, which sequentially performs cluster topology adaptation, OT merging, and cluster-decorated canonical representation and cluster-constrained alignment. 
GMaplatent inherits the advantages of AE-OT to avoid mode collapse with the same reason that uncertain margins among clusters are removed and a point or a curve in the whole connected latent space has its own semantics of category. 
The main differences include: 
(1) GMapLatent focuses on canonical latent representation and its based precise cross-domain alignment, and therefore works for cross-domain generation. 
It can also work for a single domain generation by ignoring the domain alignment procedure. We should recognize that the quality of corresponding generation depends on the capabilities of the encoder and decoder, and thus given the same set of sampled points in latent space and the same encoder-decoder baseline, GMapLatent and AE-OT for a single domain generate the same results;
(2) GMapLatent has straight boundaries of planar polygonal clusters, which help accurately and efficiently localize and detect the classes of the samples especially for those along boundaries, better than the curvy cluster boundaries in AE-OT; and
(3) GMapLatent can work for the whole domain without data cleaning and can deal with  clusters with mixtures and outliers, which respects the original data and makes model more general. 

\vspace{1mm}
\paragraph{2D latent visualization} 
We employ 2D latent representation to simplify the high-dimensional latent space alignment problem. In 2D latent space, geometric mapping and registration methodologies can be directly introduced. In this work, t-SNE representation is used to build 2D latent space; t-SNE latent code is used to establish the correspondence between the original source code and target code. From the projected target t-SNE latent code, we obtain its neighboring relationship and use that to interpolate corresponding original 150-dimensional latent codes. The interpolated high-dimensional latent code is then sent to the target's decoder for generation to guarantee the generation quality. 
In addition, the ``outliers" mentioned in our model refers to points in 2D latent space that are far from their cluster cores or occur in other clusters, rather than to true outliers in the ambient space. Our model handles all samples with the same probability measure and a uniform distribution by OT merging, which is more common and practical in real-world environments with a large number of samples. Therefore, no matter what kind of dimensionality reduction method is used, our model can still work. However, different visualization techniques (such as UMAP) may generate different clustering effects, which affects the accuracy of locating a test sample to its true category and further affects the accuracy of generation. The more the classes are separated, the better the accuracy will be. 

\vspace{1mm}
\paragraph{Advantages} 
GMapLatent has the following properties: (1) \textit{Geometrization and normalization}: it investigates the geometry and topology of the latent space of semantically labeled data and presents its intrinsic geometric representations and exact geometric alignment; (2) \textit{Generalizability and robustness}: it handles clusters with mixed and outlier data and the proposed new modules can be used as plug-ins for general domain adaptation problems; and (3) \textit{Interpretability and controllability}: it implements strict cluster constraints and canonical convex representations, providing fine-grained interpretation and control over sampling and generation.

\vspace{1mm}
\paragraph{Further improvement} 
Experiments demonstrate that the latent space layout consistency strategy used in this study is feasible and effective. Specifically, in the case of no outliers, the goal can be achieved with at most one adjustment after the initial setup; in the full domain case, using the initial setup of the specified layout works well. We still expect and will continue to explore an ideal deterministic computational design that can directly achieve consistency in merging results.

%% file: 7_conclusion.tex
\section{Conclusion and future work}
\label{sec:conclusion}
This work explores latent representations in encoder-decoder generative AI architectures and firstly proposes an interpretable and controllable cross-domain alignment and generation model that integrates diffeomorphic geometric mapping into the latent space, named GMapLatent. The process involves: 1) Converting latent spaces to canonical parameter domains through barycenter translation, optimal transport merging, and graph-constrained harmonic mapping; and 2) Performing linear constrained harmonic registration of these domains to achieve seamless diffeomorphic mapping. The interpretability and controllability of the proposed generative model lies in the use of cluster constraints and canonical latent representation, thus preserving the structures of data representations across domains. Experiments on image translation tasks demonstrate the model's efficiency, efficacy, and applicability, and exhibits the advantages comparing to the state-of-the-arts. 

In future, we will further optimize the model by connecting to other deep architectures (e.g. GAN) to improve inter-class distinguishability. As a long-term goal, we plan to delve further into the GMapLatent framework, considering more geometric variations driven by data physic and integration with large model, and exploring its broader applications for large-scale domain adaptation, knowledge transfer and multimodal information fusion in engineering and biomedicine.

%% file: main_arxiv.bbl
\begin{thebibliography}{10}

\bibitem{liu2017unsupervised}
Ming-Yu Liu, Thomas Breuel, and Jan Kautz.
\newblock Unsupervised image-to-image translation networks.
\newblock {\em Advances in Neural Information Processing Systems}, 30, 2017.

\bibitem{sainath2012auto}
Tara~N Sainath, Brian Kingsbury, and Bhuvana Ramabhadran.
\newblock Auto-encoder bottleneck features using deep belief networks.
\newblock In {\em 2012 IEEE International Conference on Acoustics, Speech and Signal Processing (ICASSP)}, pages 4153--4156. IEEE, 2012.

\bibitem{mcinnes2018umap}
Leland McInnes, John Healy, and James Melville.
\newblock Umap: Uniform manifold approximation and projection for dimension reduction.
\newblock {\em arXiv preprint, arXiv:1802.03426}, 2018.

\bibitem{van2008visualizing}
Laurens Van~der Maaten and Geoffrey Hinton.
\newblock Visualizing data using t-sne.
\newblock {\em Journal of Machine Learning Research}, 9(11), 2008.

\bibitem{zhu2017unpaired}
Jun-Yan Zhu, Taesung Park, Phillip Isola, and Alexei~A Efros.
\newblock Unpaired image-to-image translation using cycle-consistent adversarial networks.
\newblock In {\em Proceedings of the IEEE International Conference on Computer Vision}, pages 2223--2232, 2017.

\bibitem{gu2022keypoint}
Xiang Gu, Yucheng Yang, Wei Zeng, Jian Sun, and Zongben Xu.
\newblock Keypoint-guided optimal transport with applications in heterogeneous domain adaptation.
\newblock {\em Advances in Neural Information Processing Systems}, 35:14972--14985, 2022.

\bibitem{ChineseMnist}
https://www.kaggle.com/datasets/gpreda/chinese-mnist.

\bibitem{deng2012mnist}
Li~Deng.
\newblock The mnist database of handwritten digit images for machine learning research.
\newblock {\em IEEE Signal Processing Magazine}, 29(6):141--142, 2012.

\bibitem{an2019ae}
Dongsheng An, Yang Guo, Na~Lei, Zhongxuan Luo, Shing-Tung Yau, and Xianfeng Gu.
\newblock Ae-ot: A new generative model based on extended semi-discrete optimal transport.
\newblock {\em International Conference on Learning Representation}, 2020.

\bibitem{Zeng2015Graph}
Muhammad Razib, Zhong-Lin Lu, and Wei Zeng.
\newblock Structural brain mapping.
\newblock In {\em International Conference on Medical Image Computing and Computer-Assisted Intervention, Lecture Notes in Computer Science (volume 9351)}, page 760–767. Springer, 2015.

\bibitem{Zeng2018Graph}
Yi-Jun Yang, Muhammmad Razib, and Wei Zeng.
\newblock Intrinsic parameterization and registration of graph constrained surfaces.
\newblock {\em Graphical Models}, 97:30--39, 2018.

\bibitem{yang2023geolatent}
Haitao Yang, Bo~Sun, Liyan Chen, Amy Pavel, and Qixing Huang.
\newblock Geolatent: A geometric approach to latent space design for deformable shape generators.
\newblock {\em ACM Transactions on Graphics (TOG)}, 42(6):1--20, 2023.

\bibitem{muralikrishnan2022glass}
Sanjeev Muralikrishnan, Siddhartha Chaudhuri, Noam Aigerman, Vladimir~G Kim, Matthew Fisher, and Niloy~J Mitra.
\newblock Glass: Geometric latent augmentation for shape spaces.
\newblock In {\em Proceedings of the IEEE/CVF Conference on Computer Vision and Pattern Recognition}, pages 18552--18561, 2022.

\bibitem{farahani2021brief}
Abolfazl Farahani, Sahar Voghoei, Khaled Rasheed, and Hamid~R Arabnia.
\newblock A brief review of domain adaptation.
\newblock {\em Advances in Data Science and Information Engineering: proceedings from ICDATA 2020 and IKE 2020}, pages 877--894, 2021.

\bibitem{kouw2019review}
Wouter~M Kouw and Marco Loog.
\newblock A review of domain adaptation without target labels.
\newblock {\em IEEE transactions on pattern analysis and machine intelligence}, 43(3):766--785, 2019.

\bibitem{Gu2022KeypointGuidedOT}
Xiang Gu, Yucheng Yang, Wei Zeng, Jian Sun, and Zongben Xu.
\newblock Keypoint-guided optimal transport with applications in heterogeneous domain adaptation.
\newblock In {\em Neural Information Processing Systems}, 2022.

\bibitem{yang2023prototypical}
Yucheng Yang, Xiang Gu, and Jian Sun.
\newblock Prototypical partial optimal transport for universal domain adaptation.
\newblock In {\em Proceedings of the AAAI Conference on Artificial Intelligence}, volume~37, pages 10852--10860, 2023.

\bibitem{Haker2000Conformal}
Steven Haker, Sigurd~B. Angenent, Allen~R. Tannenbaum, Ron Kikinis, Guillermo Sapiro, and Michael~W. Halle.
\newblock Conformal surface parameterization for texture mapping.
\newblock {\em IEEE Transactions on Visualization and Computer Graphics}, 6:181--189, 2000.

\bibitem{Lvy2002LeastSC}
Bruno L{\'e}vy, Sylvain Petitjean, Nicolas Ray, and J{\'e}r{\^o}me Maillot.
\newblock Least squares conformal maps for automatic texture atlas generation.
\newblock {\em Proceedings of the 29th annual conference on Computer graphics and interactive techniques}, 2002.

\bibitem{gu2002computing}
Xianfeng Gu and Shing-Tung Yau.
\newblock Computing conformal structure of surfaces.
\newblock {\em arXiv preprint cs/0212043}, 2002.

\bibitem{gu2007conformal}
Xianfeng~David Gu and Shing-Tung Yau.
\newblock {\em Computational Conformal Geometry}.
\newblock Higher Education Press, 2007.

\bibitem{nehari2012conformal}
Zeev Nehari.
\newblock {\em Conformal mapping}.
\newblock Courier Corporation, 2012.

\bibitem{Zhu2003Area}
Lei Zhu, Steven Haker, and Allen Tannenbaum.
\newblock Area-preserving mappings for the visualization of medical structures.
\newblock In {\em International Conference on Medical Image Computing and Computer-Assisted Intervention (MICCAI)}, Montreal, Canada, Nov 15-18, 2003 2003.

\bibitem{Haker2004OptimalMT}
Steven Haker, Lei Zhu, Allen~R. Tannenbaum, and Sigurd~B. Angenent.
\newblock Optimal mass transport for registration and warping.
\newblock {\em International Journal of Computer Vision}, 60:225--240, 2004.

\bibitem{zhao2013area}
Xin Zhao, Zhengyu Su, Xianfeng~David Gu, Arie Kaufman, Jian Sun, Jie Gao, and Feng Luo.
\newblock Area-preservation mapping using optimal mass transport.
\newblock {\em IEEE Transactions on Visualization and Computer Graphics}, 19(12):2838--2847, 2013.

\bibitem{jost2006lectures}
J{\"u}rgen Jost.
\newblock Lectures on harmonic maps: with applications to conformal mappings and minimal surfaces.
\newblock In {\em Harmonic Mappings and Minimal Immersions: Lectures given at the 1st 1984 Session of the Centro Internationale Matematico Estivo (CIME) held at Montecatini, Italy, June 24--July 3, 1984}, pages 118--192. Springer, 2006.

\bibitem{ahlfors2006lectures}
Lars~Valerian Ahlfors.
\newblock {\em Lectures on quasiconformal mappings}, volume~38.
\newblock American Mathematical Soc., 2006.

\bibitem{zeng2009surface}
Wei Zeng, Feng Luo, S~T Yau, and Xianfeng~David Gu.
\newblock Surface quasi-conformal mapping by solving beltrami equations.
\newblock In {\em Mathematics of Surfaces XIII: 13th IMA International Conference York, UK, September 7-9, 2009 Proceedings 13}, pages 391--408. Springer, 2009.

\bibitem{zeng2012computing}
Wei Zeng, Lok~Ming Lui, Feng Luo, Tony Fan-Cheong Chan, Shing-Tung Yau, and David~Xianfeng Gu.
\newblock Computing quasiconformal maps using an auxiliary metric and discrete curvature flow.
\newblock {\em Numerische Mathematik}, 121(4):671--703, 2012.

\bibitem{floater2005surface}
Michael~S Floater and Kai Hormann.
\newblock Surface parameterization: a tutorial and survey.
\newblock {\em Advances in multiresolution for geometric modelling}, pages 157--186, 2005.

\bibitem{Lui2010OptimizationOS}
Lok~Ming Lui, Tsz~Wai Wong, Wei Zeng, Xianfeng Gu, Paul~M. Thompson, Tony~F. Chan, and Shing-Tung Yau.
\newblock Optimization of surface registrations using beltrami holomorphic flow.
\newblock {\em Journal of Scientific Computing}, 50:557--585, 2010.

\bibitem{CVPR14Point}
Wei Zeng, Lok~Ming Lui, and Xianfeng Gu.
\newblock Surface registration by optimization in constrained diffeomorphism space.
\newblock In {\em IEEE Conference on Computer Vision and Pattern Recognition}, Columbus, Ohio, USA, Jun 24-27 2014.

\bibitem{ECCV14Curve}
Wei Zeng and Yi-Jun Yang.
\newblock Surface matching and registration by landmark curve-driven canonical quasiconformal mapping.
\newblock In {\em The 13th European Conference on Computer Vision}, Zurich, Switzerland, Sep 6-12 2014.

\bibitem{yang2024diffeomorphic}
Yi-Jun Yang, Yu-Ming Zhao, Li-Qun Yang, and Wei Zeng.
\newblock Diffeomorphic registration of 3d surfaces with point and curve landmarks.
\newblock {\em Communications in Mathematics and Statistics}, 12(3):505--522, 2024.

\bibitem{su2015optimal}
Zhengyu Su, Yalin Wang, Rui Shi, Wei Zeng, Jian Sun, Feng Luo, and Xianfeng Gu.
\newblock Optimal mass transport for shape matching and comparison.
\newblock {\em IEEE Transactions on Pattern Analysis and Machine Intelligence}, 37(11):2246--2259, 2015.

\bibitem{brenier1991polar}
Yann Brenier.
\newblock Polar factorization and monotone rearrangement of vector-valued functions.
\newblock {\em Communications on Pure and Applied Mathematics}, 44(4):375--417, 1991.

\bibitem{gu2013variational}
Xianfeng Gu, Feng Luo, Jian Sun, and S-T Yau.
\newblock Variational principles for minkowski type problems, discrete optimal transport, and discrete monge-ampere equations.
\newblock {\em arXiv preprint, arXiv:1302.5472}, 2013.

\bibitem{tutte1963draw}
William~Thomas Tutte.
\newblock How to draw a graph.
\newblock {\em Proceedings of the London Mathematical Society}, 3(1):743--767, 1963.

\bibitem{floater2003mean}
Michael~S Floater.
\newblock Mean value coordinates.
\newblock {\em Computer aided geometric design}, 20(1):19--27, 2003.

\bibitem{yu2021frechet}
Yu~Yu, Weibin Zhang, and Yun Deng.
\newblock Frechet inception distance (fid) for evaluating gans.
\newblock {\em China University of Mining Technology Beijing Graduate School}, 3, 2021.

\bibitem{el2017cnn}
Ahmed El-Sawy, Hazem El-Bakry, and Mohamed Loey.
\newblock Cnn for handwritten arabic digits recognition based on lenet-5.
\newblock In {\em Proceedings of the International Conference on Advanced Intelligent Systems and Informatics 2016 2}, pages 566--575. Springer, 2017.

\bibitem{mustafa2020transformation}
Aamir Mustafa and Rafa{\l}~K Mantiuk.
\newblock Transformation consistency regularization--a semi-supervised paradigm for image-to-image translation.
\newblock In {\em Computer Vision--ECCV 2020: 16th European Conference, Glasgow, UK, August 23--28, 2020, Proceedings, Part XVIII 16}, pages 599--615. Springer, 2020.

\bibitem{korotin2019wasserstein}
Alexander Korotin, Vage Egiazarian, Arip Asadulaev, Alexander Safin, and Evgeny Burnaev.
\newblock Wasserstein-2 generative networks.
\newblock {\em arXiv preprint, arXiv:1909.13082}, 2019.

\bibitem{makkuva2020optimal}
Ashok Makkuva, Amirhossein Taghvaei, Sewoong Oh, and Jason Lee.
\newblock Optimal transport mapping via input convex neural networks.
\newblock In {\em International Conference on Machine Learning}, pages 6672--6681. PMLR, 2020.

\bibitem{choi2020stargan}
Yunjey Choi, Youngjung Uh, Jaejun Yoo, and Jung-Woo Ha.
\newblock Stargan v2: Diverse image synthesis for multiple domains.
\newblock In {\em Proceedings of the IEEE/CVF Conference on Computer Vision and Pattern Recognition}, pages 8188--8197, 2020.

\bibitem{brock2018large}
Andrew Brock.
\newblock Large scale gan training for high fidelity natural image synthesis.
\newblock {\em arXiv preprint, arXiv:1809.11096}, 2018.

\end{thebibliography}
